%% file: main.tex
\documentclass[10pt,twocolumn,letterpaper]{article}

\usepackage[pagenumbers]{cvpr} 

\definecolor{cvprblue}{rgb}{0.21,0.49,0.74}
\usepackage[pagebackref,breaklinks,colorlinks,allcolors=cvprblue]{hyperref}
\usepackage{bm}
\usepackage{amsmath}
\usepackage{graphicx}
\usepackage[linesnumbered, ruled, lined]{algorithm2e}
\usepackage{multirow}

\DeclareFixedFont{\ttb}{T1}{txtt}{bx}{n}{8} 
\DeclareFixedFont{\ttm}{T1}{txtt}{m}{n}{8}  

\usepackage{color}
\definecolor{deepblue}{rgb}{0,0,0.5}
\definecolor{deepred}{rgb}{0.6,0,0}
\definecolor{deepgreen}{rgb}{0,0.5,0}

\newcommand{\figref}[1]{Fig.~\ref{#1}}
\newcommand{\tabref}[1]{Tab.~\ref{#1}}
\newcommand{\secref}[1]{Sec.~\ref{#1}}

\def\paperID{6148} 
\def\confName{CVPR}
\def\confYear{2025}

\title{Gaussian Object Carver: Object-Compositional Gaussian Splatting \\ with Surfaces Completion}

\author{
Liu Liu\textsuperscript{*}
,
Xinjie Wang\textsuperscript{*}
, 
Jiaxiong Qiu
, 
Tianwei Lin
, 
Xiaolin Zhou
, 
Zhizhong Su\\
Horizon Robotics, Beijing, China\\
{*} Equal contribution\\
}

\begin{document}
\maketitle
\input{sec/0_abstract}

\input{sec/1_intro}

\input{sec/2_formatting}
{
    \small
    \bibliographystyle{ieeenat_fullname}
    \bibliography{main}
}

\setcounter{section}{0}  
\input{sec/X_suppl}

\end{document}

%% file: sec/0_abstract.tex
\begin{abstract}
3D scene reconstruction is a foundational problem in computer vision. Despite recent advancements in Neural Implicit Representations (NIR), existing methods often lack editability and compositional flexibility, limiting their use in scenarios requiring high interactivity and object-level manipulation. In this paper, we introduce the Gaussian Object Carver (GOC), a novel, efficient, and scalable framework for object-compositional 3D scene reconstruction. GOC leverages 3D Gaussian Splatting (GS), enriched with monocular geometry priors and multi-view geometry regularization, to achieve high-quality and flexible reconstruction. Furthermore, we propose a zero-shot Object Surface Completion (OSC) model, which uses 3D priors from 3d object data to reconstruct unobserved surfaces, ensuring object completeness even in occluded areas. Experimental results demonstrate that GOC improves reconstruction efficiency and geometric fidelity. It holds promise for advancing the practical application of digital twins in embodied AI, AR/VR, and interactive simulation environments. The code will be available at \url{https://github.com/liuliu3dv/GOC}.
\end{abstract}

%% file: sec/1_intro.tex
\section{Introduction}
\label{sec:intro}
\begin{figure}
    \centering
    \includegraphics[width=1\linewidth]{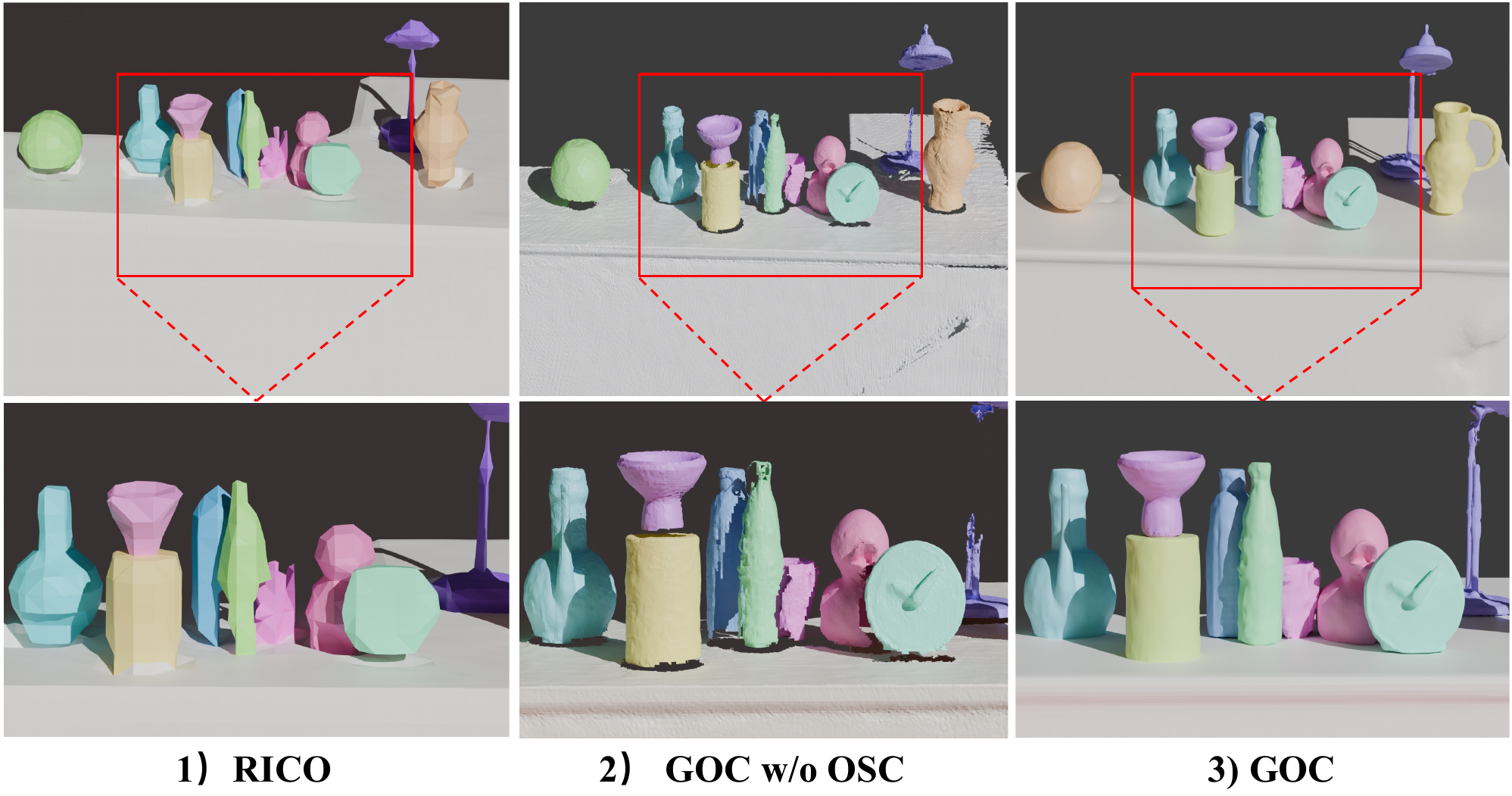}
    \vspace*{-7mm}
    \caption{Invisible Surface Completion: We introduce a novel, efficient, and scalable framework for object-compositional 3D scene reconstruction, specifically designed to complete object surfaces in occluded regions. Compared to RICO \cite{li2023rico}, GOC without OSC has better detail but suffers from surface holes. With the incorporation of OSC, our method generates watertight, separable object meshes, even in the presence of occlusions.}
    \vspace*{-5mm}
    \label{fig:enter-label}
\end{figure}

In embodied AI, collecting data from real-world environments is prohibitively expensive, making simulators a more efficient alternative. However, traditional simulators that rely on graphical assets face two main challenges: limited diversity and a domain gap between synthetic assets and real-world scenes. Recent advancements in Neural Implicit Representations (NIR) enable scalable digital twins of the real world from captured data. This capability has driven numerous downstream applications in embodied AI. For instance, in autonomous driving, neural simulators like UniSim and NeuRAD~\cite{yang2023unisim, tonderski2024neurad} enable safe, modifiable environments for effective closed-loop testing. In the field of robotics, building digital twins using NIR, often referred to as Real2Sim, shows promising potential for data collection and closed-loop training~\cite{torne2024reconciling}. However, robotics scenarios bring additional challenges, including higher-frequency physical interactions, complex occlusions, and open-set object categories.

Our goal is to design an efficient, scalable, and high-quality object-compositional scene reconstruction framework that enhances editability and interactivity, thereby expanding the applicability of NIR in embodied AI. Recent methods~\cite{wu2022object, wu2023objectsdf++, li2023rico, ni2024phyrecon} achieve object-compositional reconstruction by jointly optimizing scene geometry and segmentation. However, the training of SDF-based approaches is computationally intensive, which limits their scalability. Furthermore, in indoor scenes, complex occlusions and constrained viewpoints are common, which severely affect the quality and usability of object reconstruction. 

The motivation behind this work is to combine scene observations with data-driven priors to create digital twins from real-world log data. By leveraging observations, we can capture as much information as possible from the scene, while data-driven priors enable robust reconstruction even in cases with insufficient observations. In this paper, we introduce Gaussian Object Carver (GOC), an efficient object-compositional reconstruction framework, which is presented in \ref{fig:fig2}.We are the first to apply 3D Gaussian Splatting (3D GS) to object-compositional reconstruction, significantly improving efficiency with fast differentiable rasterization. To achieve object-separable and accurate surface reconstructions, we integrate monocular semantic and geometric priors with multi-view geometric regularization. Additionally, to address unobserved surface reconstruction, we propose a novel generative Object Surface Completion (OSC) module that completes missing regions using 3D object priors. As shown in \ref{fig:enter-label}, GOC offers better detail compared to existing methods, and can generate watertight object meshes, even in the presence of occlusions.

The contributions are summarized as follows:
\begin{itemize}[leftmargin=*] 
\item[1.] We propose Gaussian Object Carver, a novel and efficient framework that combines 3D Gaussian Splatting with a generalizable object completion model. Compared to existing methods, our approach achieves more than 10 times efficiency, and generates watertight, separable object meshes, even in scenarios involving occlusion.

\item[2.] We develop a 3D GS-based object-compositional reconstruction method incorporating monocular geometry priors and multi-view geometry regularization to improve geometric accuracy and scalability.

\item[3.] We introduce a zero-shot 3D Object Surface Completion (OSC) model, trained on a large-scale dataset, demonstrating generalizability for unseen surface completion at the object level.
\end{itemize}

%% file: sec/2_formatting.tex
\begin{figure*}[t]
\centering
\includegraphics[width=\textwidth]{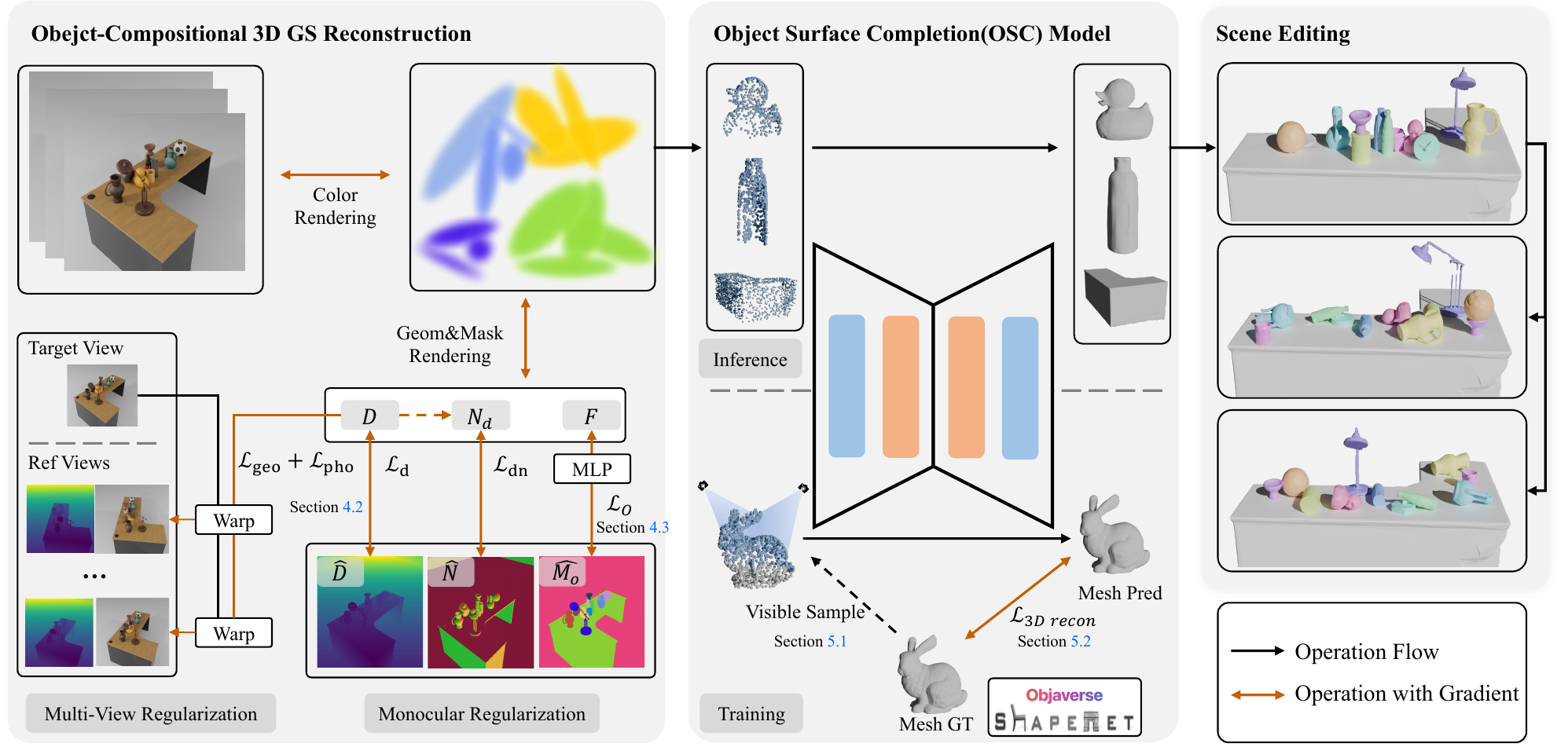}
\vspace*{-7mm}
\caption{Overview of GOC: Given multi-view images of a scene, we optimize 3D Gaussian Splatting (3D GS) to generate scene geometry and segmentation, applying regularization from both multi-view geometry and monocular priors. Next, incomplete objects from partially observed inputs are fed into the Object Completion Model (OSC), which performs zero-shot completion to fill in missing geometry and produce complete 3D shapes. Finally, this process yields watertight and separable object meshes, enabling flexible scene rearrangement and object-level manipulation.}
\vspace*{-3mm}
\label{fig:fig2}
\end{figure*}

\section{Related Work}
\label{sec:rel_work}

\subsection{3D representations and Surface Reconstruction}
Neural Radiance Fields (NeRF) \cite{mildenhall2021nerf} utilize volume rendering to create photorealistic scene representations through stable optimization. However, NeRF alone struggles with precise geometric reconstruction, leading to the development of methods that integrate geometry-based representations, such as iso-surfaces (e.g., occupancy fields \cite{mescheder2019occupancy} and Signed Distance Functions (SDF) \cite{wang2021neus, yu2022monosdf}), as well as volume density, to improve surface reconstruction fidelity. To further enhance the quality and robustness of surface reconstructions, recent approaches like MonoSDF and NeuRIS \cite{yu2022monosdf, wang2022neuris} incorporate geometric regularization from monocular models, adding constraints that help to capture fine details. Additionally, GeoNeuS\cite{Fu2022GeoNeus} introduces geometric consistency from the multi-view stereo, addressing issues of scale ambiguity and improving cross-view alignment for higher-fidelity reconstructions. These advancements have collectively enhanced the reliability of NIR for high-quality surface reconstruction.

Despite progress with NeRF and SDF approaches, optimization remains time-intensive. Recently, 3D Gaussian Splatting (3D GS)~\cite{ye2023gaussian} has redefined efficiency in 3D reconstruction, offering high-quality, fast rendering through differentiable rasterization of 3D Gaussians. 
Achieving geometric accuracy and meshable surfaces from Gaussian primitives is increasingly critical for 3D scene understanding. Recent works~\cite{guedon2024sugar, huang20242d,chen2024pgsr} aim to refine geometric precision and mesh generation from these representations. Likewise, DN-Splatter~\cite{turkulainen2024dn} demonstrates that depth and normal priors can significantly enhance the training of 3D Gaussian splatters, leading to higher fidelity in reconstruction and meshing. A crucial aspect of compositional scene reconstruction is high-quality 3D segmentation. Recent approaches~\cite{gaussian_grouping,wu2024opengaussian,zhou2024feature,qu2024goi} combine 2D scene understanding with 3D Gaussians, enabling real-time, editable 3D scene representations that address the computational inefficiencies of NeRF-based methods. By using consistent 2D masks across views, Gaussian Grouping achieves enhanced segmentation quality and computational efficiency compared to NeRF-based techniques.

In this work, to improve training efficiency, we introduce 3D Gaussian Splatting into object-compositional scene reconstruction. These enable our framework to achieve state-of-the-art quality and efficiency in object-compositional reconstruction.

\subsection{Compositional Scene Reconstruction}

Recent methods~\cite{wu2022object, wu2023objectsdf++} achieve object-compositional reconstruction by disentangling objects within a scene. Building upon ObjectSDF++, additional work \cite{wu2025clusteringsdf} addresses the dependency on annotations, while \cite{ni2024phyrecon} introduces physically differentiable constraints, collectively enhancing the practicality of SDF-based object-compositional reconstruction. However, in indoor scenes, complex occlusions and restricted viewpoints are common, severely affecting the quality and usability of object reconstruction. Methods like \cite{li2023rico, hassena2024objectcarver} leverage scene geometry priors, such as background smoothness, object compactness, and object-background relationships, through the designed SDF regularization. Relying on manually designed regularization terms has limitations, as it cannot address challenging scenes with complex occlusions and restricted viewpoints.



Our approach combines scene observations with data-driven priors to address incomplete object observations. First, we leverage a 3D GS-based object-compositional reconstruction to extract observed geometry. Then, we complete the unobserved surfaces using 3D object priors, enabling more accurate and realistic scene reconstructions.

\subsection{3D Object Completion}
For 3D shape completion, methods like PCN  \cite{yuan2018pcn}, TopNet  \cite{shu20193d}, and GRNet \cite{xie2020grnet} address the task by transforming partial point clouds into complete ones. However, to obtain a watertight mesh, these approaches require additional surface reconstruction algorithms (e.g. traditional \cite{kazhdan2006poisson, newcombe2011kinectfusion_tsdf} or neural kernel-based surface reconstruction \cite{huang2023neural, williams2022neural, williams2021neural}) to post-process the completed point clouds. Newer methods, such as PatchComplete  \cite{rao2022patchcomplete} and DiffComplete \cite{chu2024diffcomplete}, focus on directly completing missing signed distance fields (SDFs), resulting in a complete SDFs for a watertight shape. However, acquiring partial SDFs from captured or reconstructed point clouds is challenging, which limits their practicality in real-world applications where SDFs may not be readily available. Transformer-based models like ShapeFormer \cite{yan2022shapeformer} enhance global feature learning by taking partial points as input and decoding them into a local deep implicit function, from which a mesh can be extracted via methods such as Marching Cubes \cite{we1987marching}. Nevertheless, these models often rely on small, category-specific datasets, restricting their ability to generalize to unseen objects and complex scenes. 

Methods like \cite{xu2024instantmesh, wu2024unique3d, liu2024one, liu2024one2} use a single-view image to generate the object mesh and have trained on a large amount of data. Additionally, works like OccNet \cite{mescheder2019occupancy}, 3dshape2vecset \cite{zhang20233dshape2vecset}, GEM3D \cite{petrov2024gem3d} and CLAY \cite{zhang2024clay} leverage additional modalities, such as point clouds or text prompt as conditions for diffusion models. However, these methods typically focus on individual object generation and struggle with object separation from the whole scene. 

Our approach introduces a unified framework that combines scene reconstruction, instance segmentation, and object completion. This framework not only reconstructs and completes individual watertight meshes for each object but also maintains the original geometric structure, achieving great generalization across diverse shape collections.

\section{Overall Framework}
\label{sec:framework}

As shown in ~\figref{fig:fig2}, Gaussian Object Carver (GOC) is 3D GS-based object-compositional reconstruction framework. Given multi-view images of a scene as input, GOC efficiently generates separable object meshes, enabling flexible scene editing and object-level manipulation. To address the challenge of compositional reconstruction, the framework consists of two primary modules. The first is a 3D GS-based object-compositional reconstruction method, which is detailed in \secref{sec:method_3dgs_oc}. In this section, we describe our approach and the design of regularization techniques for optimization.
The second module is an general Object Completion Model (OSC), which is specifically designed to handle incomplete or occluded objects. Leveraging object priors, OSC generates complete geometric reconstructions of partially observed objects. Further details of the OSC are provided in \secref{sec:method_oscm}, where we describe its architecture and functionality in greater depth.




\section{3D GS-based Compositional Reconstruction}
\label{sec:method_3dgs_oc}
This section is organized into four parts. Firstly, we review the preliminary concepts of 3D GS in Section~\ref{subsec:reliminary_3dgs}. Then, in Section~\ref{subsec:geom_reg_3dgs}, we present the regularization of geometry. Next, we discuss the rendering and regularization of segmentation in Section~\ref{subsec:seg_reg_3dgs}. Finally, in Section~\ref{subsec:implementaion_3dgs}, we describe the optimization procedure. More implementaion details are explained in supplementary materials.

\subsection{Preliminary}
\label{subsec:reliminary_3dgs}
Our work builds on 3D Gaussian Splatting~\cite{kerbl20233d_3dgs}, and the scene is explicitly represented by numerous differentiable 3D Gaussian primitives ${\bm{G}}$. Each primitive is parameterized by a mean $\mu \in \mathbb{R}^3$, a covariance matrix $\Sigma \in \mathbb{R}^{3 \times 3}$, which is decomposed into a scaling vector $s \in \mathbb{R}^3$ and a rotation quaternion $q \in \mathbb{R}^4$, along with opacity $o \in \mathbb{R}$ and color $c \in \mathbb{R}^3$, the latter represented using spherical harmonics.

With patchwise parallelization, 3DGS achieves efficient alpha-blending for rendering and training. For each camera view, after 3D Gaussian primitives are projected as 2D space and sorted by z-buffer. Then the color C of a pixel could be computed by volumetric rendering\cite{mildenhall2021nerf} using front-to-back depth order. The composite pixel-wise color \( C \) and alpha \( A \) are given by:
\begin{equation}
\setlength{\abovedisplayskip}{3pt}
\begin{aligned}
C &= \sum_{i \in \mathcal{N}} c_i \alpha_i T_i,\\
A &= \sum_{i \in \mathcal{N}}\alpha_i T_i, 
\end{aligned}
\label{eq:volumn_rendering}
\setlength{\belowdisplayskip}{3pt}
\end{equation}

Where $T_i = \prod_{j=1}^{i-1} (1 - \alpha_j)$, N is the set of sorted Gaussians on the ray of rendered pixel, and ${T_i}$ is the transmittance, defined as the product of opacity values of previous Gaussians overlapping the same pixel.

\subsection{Geometry Regularization}
\label{subsec:geom_reg_3dgs}

For depth rendering, to eliminate the transparency impact on depth rendering, the rendered depth $D$ needs to be alpha-normalized based on pixel alpha and is computed as:

\begin{equation}
\setlength{\abovedisplayskip}{3pt}
\begin{aligned}
    D &= \sum_{i \in \mathcal{N}} d_i \alpha_i \prod_{j=1}^{i-1} (1 - \alpha_j)/A
\end{aligned}
\label{eq:deptg_rendering}
\setlength{\belowdisplayskip}{3pt}
\end{equation}

where $d$ is the distance between the 3D Gaussian center and the camera center.

\paragraph{Monocular Geometry Regularization}
Surface reconstruction in complex indoor scenes is inherently challenging due to the lack of texture. To address this, we draw inspiration from \cite{yu2022monosdf,turkulainen2024dn} and propose incorporating monocular priors, specifically normal $\hat{N}$ and depth $\hat{D}$, into the reconstruction process.

First, the rendered depth$D$ can be directly constrained through depth prior $\hat{D}$ by:
\begin{equation}
\setlength{\belowdisplayskip}{3pt}
\mathcal{L}_{d} = \sum_{i,j}|D-\hat{D}|
\label{eq:depth_reg_loss}
\setlength{\belowdisplayskip}{3pt}
\end{equation}
By deriving the depth result, we can compute the normal from rendered depth $N_d$. We then leverage $\hat{N}$ to regularize this result. To mitigate the impact of transparency, we use $\alpha$ to weight the normal loss:
\begin{equation}
\setlength{\belowdisplayskip}{3pt}
\mathcal{L}_{dn} = \sum_{i,j}\alpha(1-N_d^T\hat{N})
\label{eq:depth2normal_loss}
\setlength{\belowdisplayskip}{3pt}
\end{equation}

\paragraph{Multi-View Geometry Regularization} Monocular depth estimation often suffers from ambiguities, leading to inconsistencies across views and degrading surface reconstruction quality. To address this, we integrate multi-view geometry regularization into optimization, inspired by prior work in multi-view geometry (\cite{monodepth17, Fu2022GeoNeus}).

We introduce a photometric reprojection loss inspired by self-supervised depth estimation~\cite{monodepth17}. Unlike methods requiring optical flow priors or multi-plane projections, our approach is computationally efficient and needs no preprocessing. With monocular geometry priors providing reasonably accurate depth estimates, this loss also avoids local optima:

\begin{equation} \setlength{\abovedisplayskip}{3pt} \mathcal{L}_{\text{pho}} = \frac{1}{N} \sum_{i,j} \lambda \frac{1 - \text{SSIM}(C_{ij}, \tilde{C}_{ij})}{2} + (1 - \lambda) \left| C_{ij} - \tilde{C}_{ij} \right|, \setlength{\belowdisplayskip}{3pt} \end{equation}

Where $\lambda = 0.85$, $\tilde{C}_{ij}$ are the colors from the reference frame projected using rendered depth $D$, and $C{ij}$ are the colors in the target frame.

Photometric reprojection constraints may struggle in low-texture or overexposed regions. To enhance 3D consistency, we adopt the geometry reprojection consistency loss $\mathcal{L}_{\text{geom}}$ from \cite{chen2024pgsr}. This term enforces depth alignment across viewpoints by computing the circular projection error between depth maps from the reference and target frames.

\subsection{Segmentatin Regularization}
\label{subsec:seg_reg_3dgs}

For object segmentation, inspired by \cite{gaussian_grouping}, we first give 3D Gaussian a group of learnable features $f$ encoded semantic information and render the feature for each pixel through alpha-blending. Similar to color rendering, the rendered semantic features $\bm{F}$ can be produced by:
\begin{equation}
\setlength{\abovedisplayskip}{3pt}
\begin{aligned}
F &= \sum_{i \in N}\bm{f}_i\alpha_i\prod_{j=1}^{i-1}(1-\alpha_j)
\end{aligned}
\label{eq:rendering}
\setlength{\belowdisplayskip}{3pt}
\end{equation}
where $\bm{f}_i$ is the semantic feature of each Gaussian primitive. 

Then, we use a multilayer perceptron network (MLP) and softmax to get the classification from the rendered semantic feature ${F}$, thus obtaining an instance mask for each pixel${M_o}$. We use a cross entropy loss $\mathcal{L}_o$ between instance mask GT ${\hat{M_o}}$ and ${M_o}$.

\setlength{\belowdisplayskip}{3pt}

\subsection{Optimization}
\label{subsec:implementaion_3dgs}
\label{sec:gs_opt}
We adopt the photometric loss $\mathcal{L}_c$ from vanilla 3DGS~\cite{kerbl20233d_3dgs}. 
All loss functions are simultaneously optimized by training from scratch. The total loss function $\mathcal{L}$ can be defined as:
\begin{equation}
\label{eq:total_loss}
\setlength{\abovedisplayskip}{3pt}
\mathcal{L} = \mathcal{L}_c + \lambda_{d}\mathcal{L}_{d} +\lambda_{dn}{L}_{dn}+\lambda_{geo}\mathcal{L}_{geo} +  \lambda_{pho}\mathcal{L}_{pho} + \lambda_{o}\mathcal{L}_{o}
\end{equation}
In our experiments, ${\lambda}_d=0.3$, ${\lambda}_n=0.1$, ${\lambda}_{pho}=0.3$, ${\lambda}_{geo}=0.3$, ${\lambda}_{o}=0.1$.
And we use the training strategy in \cite{3d_gs_mcmc}, because we observed that the 3DGS strategy \cite{kerbl20233d_3dgs} is sensitive to initialization and hyperparameter settings. 


\section{Obejct Surfaces Completion Model}
\label{sec:method_oscm}

We propose a general Object Surface Completion (OSC) model, designed to recover complete, watertight meshes from sparse or partial point clouds. The objective of OSC is to reconstruct meshes that closely resemble the original geometric structure of the object, rather than the diversity in generated meshes. To achieve this, we adopt a lightweight VAE \cite{Kingma2013AutoEncodingVB} framework. OSC demonstrates strong generalization capability, enabling surface completion and reconstruction of objects with arbitrary geometries without fine-tuning in different domains. It supports inputs from incomplete point clouds collected by real sensors, depth recovery, or 3DGS \cite{kerbl20233d_3dgs} and NeRF \cite{mildenhall2021nerf} based reconstructions.

This section is organized into three parts. First, Sections~\ref{sec:model} and \ref{sec:SOC_Losses} introduce the model details and the loss design of the OSC model. Next, Section~\ref{sec:train_gocm_detail} provides detailed insights into the training process of the OSC model.

\subsection{Model Details}
\label{sec:model}

The OSC model encodes the geometric structure of the input point cloud into an implicit latent space through the Surface Points Encoder. Using grid query points and the embedding of the encoded point cloud, the surface completion decoder then outputs the occupancy probability, the likelihood that each point lies within the object’s surface, for each query point. Finally, the Marching Cubes algorithm \cite{we1987marching} is applied to extract the surface mesh.

\begin{figure}
    \centering
    \includegraphics[width=\linewidth]{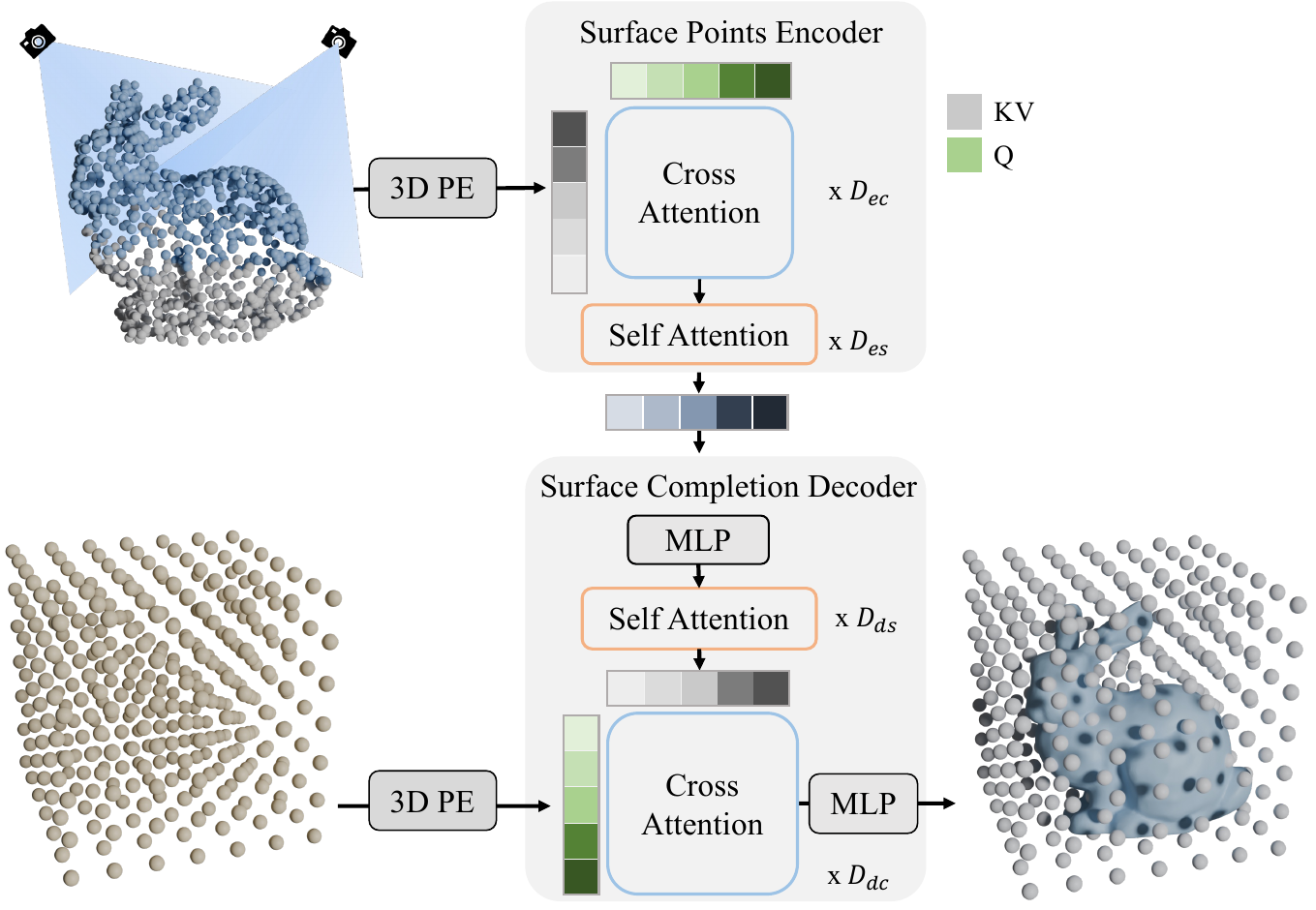}
    \vspace*{-7mm}
    \caption{ Illustration of OSC for surface reconstruction from sparse or incomplete point clouds. During training, points are uniformly sampled from the object’s surface mesh, filtered by specified virtual cameras, and encoded into embedding in latent space via an encoder. The decoder predicts the occupancy probability of each query point in a predefined 3D grid with the embedding. The reconstructed complete mesh is obtained by extracting the isosurface from the occupancy field. }
    \label{fig:osc_model}
    \vspace*{-3mm}
\end{figure}

\paragraph{Masking}
To enable the model to recover complete watertight surfaces from incomplete point clouds, we extend the MAE \cite{he2022masked} masking strategy into 3D space. Specifically, during model training, we uniformly sample points from the mesh surface and project the points into the camera's pixel coordinates using the intrinsic and extrinsic parameters of the virtual camera. Points that do not fall onto the imaging plane are masked out. Considering depth occlusions between points, non-visible points are also masked. The retained point set is denoted as $\mathbf{P_s}$. Gaussian noise is added to $\mathbf{P_s}$ to augment data with slight random perturbations.

\paragraph{Surface points Encoder}

 We apply Farthest Point Sampling (FPS) \cite{qi2017pointnet++} to extract $\mathbf{P_s}$ core structure into the dimensions $\mathbb{R}^{M \times 3}$, which is encoded into the latent space through Fourier Positional Encodings (FPE) \cite{mildenhall2021nerf} to serve as learnable queries. The point set $\mathbf{P_s}$ is also encoded with FPE as Key-Value (KV). After passing through cross-attention layers with depth $D_{ec}$, followed by self-attention layers with depth $D_{es}$, we obtain the embedding representation $\mathbf{E} \in \mathbb{R}^{M \times D}$ of $\mathbf{P_s}$ in the latent space, where $M = 2048$, $D_{ec} = 10$, $D_{es} = 10$, and $D = 16$.

\begin{equation}
\label{eq:vae_enc}
\setlength{\abovedisplayskip}{1pt}
\scalebox{0.84}{$
    \mathbf{E} = \operatorname{SelfAttns} \left( \operatorname{CrossAttns} \left( \operatorname{FPE} \left( \operatorname{FPS} \left( \mathbf{P}_s \right) \right), \operatorname{FPE} \left( \mathbf{P}_s \right) \right) \right)
$}
\end{equation}
\setlength{\belowdisplayskip}{1pt}

By applying KL regularization, we constrain $\mathbf{E}$ to a Gaussian distribution, ensuring that similar data have similar positions in the latent space, facilitating the generation of continuous data as in \cite{Kingma2013AutoEncodingVB}. Specifically, two MLP layers are used to learn the Gaussian distribution’s mean $\mu_i$ and variance $\sigma_i^2$ from $\mathbf{E} \in \mathbb{R}^{M \times D}$. The regularization loss $\mathcal{L}_{\text{KL}}$ is applied to enforce this constraint:

\begin{equation}
\label{eq:vae_kl_loss}
\setlength{\abovedisplayskip}{3pt}
\mathcal{L}_{\text{KL}} = \frac{1}{2} \sum_{i=1}^D \left( \sigma_i^2 + \mu_i^2 - 1 - \log \sigma_i^2 \right)
\setlength{\belowdisplayskip}{3pt}
\end{equation}

A sampled latent $\mathbf{E_s} \in \mathbb{R}^{M \times D}$ is then drawn from the distribution and used as input to the subsequent decoder.

\paragraph{Surface Completion Decoder}

Unlike MAE \cite{he2022masked}, which compresses and reconstructs images, our surface completion decoder takes as input a set of initialized query points, $\mathbf{Q}_{g} \in \mathbb{R}^{k \times 3}$, along with the latent space encoding $\mathbf{E_s}$ of the surface point cloud. The decoder outputs an occupancy probability $\mathcal{\hat{O}}_s(\mathbf{Q}_{g}) \in \mathbb{R}^{k}$ for each query point similar to methods in \cite{mescheder2019occupancy, zhang20233dshape2vecset}:

\begin{equation}
\label{eq:vae_dec}
\setlength{\abovedisplayskip}{3pt}
\scalebox{0.85}{$
    \mathcal{\hat{O}}_s(\mathbf{Q}_g) = \operatorname{MLP} \left( \operatorname{CrossAttns} \left( \operatorname{FPE}(\mathbf{Q}_g), \operatorname{SelfAttns} \left( \mathbf{E}_s \right) \right) \right)
$}
\setlength{\belowdisplayskip}{3pt}
\end{equation}

\subsection{Optimization}
\label{sec:SOC_Losses}

Directly supervising the decoder with the ground truth $\mathcal{O} (\mathbf{Q}g) \in \{0, 1\}$ often results in noticeable artifacts, such as streaks and voxel-like patterns on the reconstructed object surface. Inspired by label smoothing \cite{muller2019does}, we set the values of $\mathcal{O} (\mathbf{Q}g)$ near the object’s surface by utilizing a signed distance field (SDF) and a threshold $T_{\text{iso}}$ to map values within the range of 0 to 1. This approach enhances reconstruction precision and smoothness around the surface as shown in supplementary materials. The threshold $T_{\text{iso}}$ is set to 1/128:

\begin{equation}
\label{eq:vae_smoothing_thresh}
\setlength{\abovedisplayskip}{3pt}
    \text{smooth}(\mathcal{O} (\mathbf{Q}_g)) = 0.5 \cdot \mathbf{1} - 0.5 \times \frac{\text{SDF}(\mathbf{Q}g)}{T_{\text{iso}}}
\end{equation}
\setlength{\belowdisplayskip}{3pt}

\begin{equation}
\label{eq:vae_smoothing}
\setlength{\abovedisplayskip}{3pt}
\setlength{\belowdisplayskip}{3pt}
\scalebox{0.83}{$
    \mathcal{O}_s(\mathbf{Q}_g) = 
    \begin{cases} 
        \mathbf{0}, & \text{if } \text{SDF}(\mathbf{Q}_g) > T_{\text{iso}} \\ 
        \text{smooth}(\mathcal{O} (\mathbf{Q}g)), & \text{if } -T_{\text{iso}} \leq \text{SDF}(\mathbf{Q}_g) \leq T_{\text{iso}} \\ 
        \mathbf{1}, & \text{otherwise}
    \end{cases}
$}
\end{equation}
\setlength{\belowdisplayskip}{3pt}

We optimize the model by using a binary cross-entropy loss to minimize the distribution difference between the predicted occupancy probability and the ground truth, similar to \cite{mescheder2019occupancy, zhang20233dshape2vecset},  $\theta$ represents the learnable parameters of the model $f_\theta$:

\begin{equation}
\label{eq:vae_loss_bce}
\setlength{\abovedisplayskip}{3pt}
\mathcal{L}_{\text{BCE}}(\theta) = \mathbb{E} \left[ \operatorname{BCE}(f_\theta(\mathbf P_s, \mathbf Q_g), \mathcal O_s(\mathbf Q_g)) \right]
\end{equation}
\setlength{\belowdisplayskip}{3pt}

The model $f_\theta$ outputs the occupancy probability for each query point, and the final mesh is formed by applying a predefined isosurface threshold $T_b$ using the Marching Cubes algorithm \cite{we1987marching}. However, optimizing solely with  $\mathcal{L}_{\text{BCE}}$  does not guarantee a clear surface boundary under threshold  $T_b$, which may lead to incomplete mesh generation as shown in supplementary materials. To address this, we introduce  $\mathcal{L}_{\text{IoU}}$, which converts each query point’s occupancy probability to an actual occupancy state based on the threshold  $T_b$. The Intersection over Union (IoU) is then computed with the original ground truth  $\mathcal{O} (\mathbf{Q}_g) \in \{0, 1\}$  to achieve clear boundaries.  $T_b$  is set to 0.3 during both training and inference:

\begin{equation}
\label{eq:vae_loss_iou_thresh}
\setlength{\abovedisplayskip}{3pt}
\mathcal{\hat{O}}(\mathbf{Q}{g}) = \begin{cases}
\mathbf{1}, & \text{if } \mathcal{\hat{O}}_s(\mathbf{Q}_{g}) > T_b \\
\mathbf{0}, & \text{otherwise}
\end{cases}
\end{equation}
\setlength{\belowdisplayskip}{3pt}

\begin{equation}
\label{eq:vae_loss_iou}
\setlength{\abovedisplayskip}{3pt}
\mathcal{L}_{\text{IoU}}(\theta) = \mathbb{E} \left[ 1 - \operatorname{IoU}\left(\hat{\mathcal{O}}(\mathbf{Q}_g), \mathcal{O}(\mathbf{Q}_g)\right) \right]
\end{equation}
\setlength{\belowdisplayskip}{3pt}

The total loss can be written as below, where $\lambda_{\text{BCE}}$, $\lambda_{\text{IoU}}$, and $\lambda_{\text{KL}}$ are set to 1.0, 0.01, and 0.0001 respectively:

\begin{equation}
\label{eq:vae_total_loss}
\setlength{\abovedisplayskip}{3pt}
\mathcal{L}(\theta) = \lambda_{\text{BCE}} \mathcal{L}_{\text{BCE}} + \lambda_{\text{IoU}} \mathcal{L}_{\text{IoU}} + \lambda_{\text{KL}} \mathcal{L}_{\text{KL}}
\end{equation}
\setlength{\belowdisplayskip}{3pt}



\subsection{Implementation Details}
\label{sec:train_gocm_detail}
The OSC training dataset consists of meshes tagged as trainset from the ShapeNet Core v2 dataset \cite{chang2015shapenet} and a diverse selection of meshes from Objaverse \cite{deitke2023objaverse}. Meshes from Objaverse were filtered to exclude those with very few faces or vertices and any that contained multiple disconnected objects. Additionally, low-quality meshes, as indicated by the annotations from \cite{zuo2024sparse3d}, were excluded. Each mesh was converted to a watertight form using TSDF \cite{newcombe2011kinectfusion_tsdf}. We then used Open3D \cite{zhou2018open3d} to sample surface points, query points, and calculate SDF values for each query point as ground truth for training. After discarding meshes for which TSDF or SDF calculations failed, the resulting dataset included a curated set of around 400,000 high-quality diverse meshes.

The training was conducted with the AdamW optimizer, using a learning rate of 1e-4 and a batch size of 8, utilizing 32 NVIDIA 4090 GPUs over 4 days.

\section{Experiments}
\label{sec:exp}
\begin{table*}[ht]
\centering
\caption{Comparison of Methods on Object and Scene Reconstruction under \textbf{Full Observation}. GOC achieves the best geometric accuracy and completeness while being highly efficient, requiring only $5\%$ of the time consumed by current state-of-the-art methods. The top-performing metrics are \textbf{highlighted}.}
\vspace*{-3mm}
\label{tab:scene_obj_comparison}
\resizebox{\linewidth}{!}{%
\begin{tabular}{lcccccccccc}
\toprule
\multirow{2}{*}{Method} & \multirow{2}{*}{Time$~\downarrow$} & \multicolumn{4}{c}{Object Recon} & \multicolumn{4}{c}{Scene Recon} \\
\cmidrule(lr){3-6} \cmidrule(lr){7-10}
& & Accuracy$~\downarrow$ & Completion$~\downarrow$ & CD$~\downarrow$ & F-score$~\uparrow$ & Accuracy$~\downarrow$ & Completion$~\downarrow$ & CD$~\downarrow$ & F-score$~\uparrow$ \\
\midrule
ObjectSDF++(MLP) \cite{wu2023objectsdf++} & 21h 26min & 0.0232 & 0.0511 & 0.0371 & 0.8741 & 0.0203 & \textbf{0.0252} & 0.0240 & 0.9164 \\
RICO \cite{li2023rico} & 17h 59min & 0.0203 & 0.0629 & 0.0416 & 0.8429 & 0.0248 & 0.0354 & 0.0330 & 0.8642 \\
GOC w/o OSC & \textbf{1h 7min} & \textbf{0.0045} & 0.0543 & 0.0294 & 0.9124 & \textbf{0.0136} & 0.0271 & \textbf{0.0211} & \textbf{0.9570} \\
GOC w ShapeFormer \cite{yan2022shapeformer} & 1h 11min & 0.0239 & 0.0689 & 0.0464 & 0.7875 & - & - & - & - \\
GOC & 1h 9min & 0.0062 & \textbf{0.0501} & \textbf{0.0282} & \textbf{0.9228} & - & - & - & - \\
\bottomrule
\end{tabular}%
}
\vspace*{-3mm}
\end{table*}

\begin{table}[ht]
\centering
\caption{Comparison of Methods on Object and Scene Reconstruction under \textbf{Sparse Observation}. The best metrics are \textbf{highlighted}.}
\vspace*{-3mm}
\label{tab:obj_comparison_sparse}
\resizebox{\linewidth}{!}{%
\begin{tabular}{lcccc}
\toprule
Method & Accuracy↓ & Completion↓ & CD↓ & F-score↑ \\
\midrule
ObjectSDF++(MLP) \cite{wu2023objectsdf++} & 0.0140 & 0.0654 & 0.0397 & 0.8749 \\
RICO \cite{li2023rico} & 0.0177 & 0.0635 & 0.0406 & 0.8354 \\
GOC w/o OSC & \textbf{0.0038} & 0.0677 & 0.0357 & 0.8682 \\
GOC w ShapeFormer \cite{yan2022shapeformer} & 0.0201 & 0.0802 & 0.0502 & 0.7835 \\
GOC & 0.0073 & \textbf{0.0575} & \textbf{0.0324} & \textbf{0.9033} \\
\bottomrule
\end{tabular}%
}
\vspace*{-3mm}
\end{table}

In this section, we first introduce our experimental setup. Then, in Section~\ref{subsec:exp_syn_scenes} and Section~\ref{subsec:exp_real_scenes}, we compare our framework with state-of-the-art methods to evaluate surface reconstruction on both synthetic and real datasets. Finally, we present ablation studies in Section~\ref{subsec:exp_ablation_study}.

\subsection{Settings}
\label{subsec:exp_setting}

\paragraph{Datasets}
In our experiments, we used two public datasets, ShapeNet \cite{chang2015shapenet} and ScanNet \cite{dai2017scannet}, as well as a custom synthetic dataset. ShapeNet, a large-scale and richly annotated shape repository represented by 3D CAD models, was employed to evaluate the surface reconstruction quality of the OSC model with complete point cloud inputs. However, our primary focus was to assess the geometric quality of each object after reconstructing the scene and completing the segmentation of all objects. Since ShapeNet consists of near-perfect individual CAD models, it is unsuitable for evaluating surface reconstruction and completion on incomplete point clouds. Similarly, ScanNet features incomplete ground truth meshes, with missing object surfaces in unobserved views, making it unsuitable for evaluating the quality of full-object completion. Consequently, a custom synthetic dataset was essential for our experiments. We created five synthetic indoor scenes, each containing approximately ten fully detailed 3D assets from BlenderKit \cite{BlenderKit}. We manually configured camera paths around each scene, rendering 170 RGB-D images along with instance masks using Blender \cite{Blender} to create a \textbf{full observation} dataset. Additionally, a \textbf{sparse observation} dataset was generated by sampling 30\% of the viewpoints, resulting in 50 images, to simulate a more challenging scenario where handheld data capture provides only partial views of object surfaces.

\paragraph{Metrics} For scene reconstruction performance, we report Chamfer Distance(CD), F-score, and normal consistency(NC) for evaluation on ScanNet. For synthetic scenes, we separate the metrics into two aspects: scene reconstruction and object completion. For object reconstruction evaluation on ShapeNet, we use Intersection over Union (IoU) as an additional metric.

\paragraph{Baselines} For the object reconstruction, segmentation, and completion task, we selected ObjSDF++\cite{wu2023objectsdf++} and RICO\cite{li2023rico} as baseline methods. We report the performance metrics separately for both the reconstruction phase and the object completion phase. For object completion, we also compare our method with ShapeFormer\cite{yan2022shapeformer}, which serves as the completion network for partial point cloud inputs. Additionally, we evaluated our OSC model on the ShapeNet test set for surface reconstruction quality with complete point cloud inputs, comparing its performance to state-of-the-art methods such as 3D2VS \cite{zhang20233dshape2vecset} and IF-Net \cite{chibane2020implicit}.

\subsection{Reconstruction in Synthetic Scenes}
\label{subsec:exp_syn_scenes}

To align with the settings of ObjectSDF++ \cite{wu2023objectsdf++} and RICO \cite{li2023rico}, all images were downsampled to a resolution of 384x384. Both ObjectSDF++ and RICO were trained for up to 3000 epochs until convergence, while our GOC’s 3D GS reconstruction was trained for 30,000 steps. For testing, sampled segmented point clouds of each reconstructed object were used as inputs to the OSC model. Pre-trained on ShapeNet \cite{chang2015shapenet} and Objaverse \cite{deitke2023objaverse}, OSC required no additional fine-tuning to generalize to the domain of reconstructed point clouds, and was used only for inference.

\begin{figure}
    \centering
    \includegraphics[width=\linewidth]{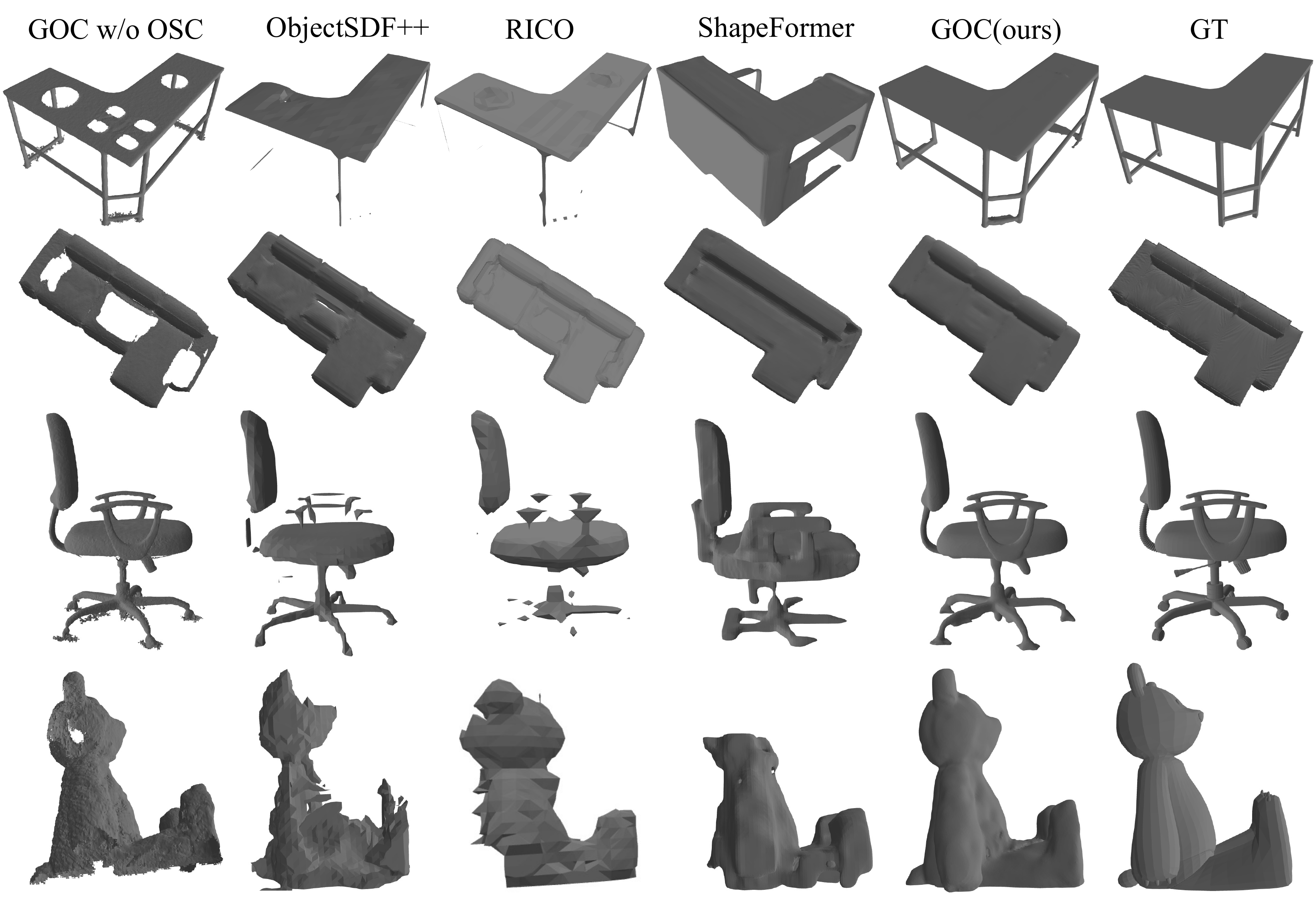}
    \vspace*{-7mm}
    \caption{Qualitative comparison of surface reconstruction and completion quality across different methods on objects from the synthetic dataset. Zoom in for details.}
    \label{fig:demo}
    \vspace*{-7mm}
\end{figure}

As shown in Table~\ref{tab:scene_obj_comparison}, our method (GOC w/o OSC, where “w/o OSC” refers to reconstruction without completion) achieves state-of-the-art performance in scene reconstruction. Compared to recent approaches like ObjectSDF++ and RICO, GOC achieves higher accuracy. However, the Completion metric reveals that our reconstruction completeness is slightly lower than ObjectSDF++. We compare the object reconstruction matrics after integrating the OSC model, which completes the segmented and sampled point clouds of reconstructed objects, we achieve a marked improvement in the completion metric, resulting in the best overall performance. When comparing our completed point cloud reconstruction with ShapeFormer, another surface completion method based on point cloud input, GOC demonstrates significant advantages in both CD and F-score, see the qualitative comparison across different methods in Figure~\ref{fig:demo}. Notably, GOC is highly efficient, consuming only $5\%$ of the time required by the current state-of-the-art methods. On average, it takes just 1 hour and 9 minutes to reconstruct an entire scene and complete all objects within it.

We assessed the geometric accuracy and completeness of object completion and reconstruction under the challenging Sparse Observation setting, as presented in Table~\ref{tab:obj_comparison_sparse}. Although our accuracy decreased slightly after applying completion (GOC w/o OSC), from 0.0073 to 0.0038, the completeness of object reconstruction improved significantly, from 0.0677 to 0.0575. Compared to other methods, our approach achieves superior performance in both Chamfer Distance and F-score.

\subsection{Reconstruction in Real-world Scenes}
\label{subsec:exp_real_scenes}
For indoor surface reconstruction, we conduct comparisons on widely used real-world datasets \cite{dai2017scannet}. The results are reported in \tabref{tab:scene_obj_comparison_scannet}. Since the ground truth data also exhibits missing occluded regions in real-world datasets, we evaluate reconstruction performance without including the Object Surface Completion (OSC) model for comparison in this experiment.

\begin{table}[ht]
\centering
\caption{Comparison of Methods on Object and Scene Reconstruction on Scannet\cite{dai2017scannet}. The top-performing metrics are \textbf{highlighted} and second-performing metrics are \underline{highlighted}. }
\label{tab:scene_obj_comparison_scannet}
\vspace*{-3mm}
\resizebox{\linewidth}{!}{%
\begin{tabular}{lcccccccc}
\toprule
\multirow{2}{*}{Method} & \multirow{2}{*}{Image size} &
\multicolumn{3}{c}{Object Recon} & \multicolumn{3}{c}{Scene Recon} \\
\cmidrule(lr){3-5} \cmidrule(lr){6-8}

& &CD$~\downarrow$ & F-score$~\uparrow$ & NC$~\uparrow$
&CD$~\downarrow$ & F-score$~\uparrow$ & NC$~\uparrow$ \\
\midrule

MonoSDF \cite{yu2022monosdf}  &384x384 & -  & -  & -  & 0.0897 & 0.6030 & 0.844 \\
RICO \cite{li2023rico}  &384x384 & 0.0929  & 0.7310 & 0.7944 & 0.0892 & 0.6144 & 0.8458 \\
ObjectSDF++ \cite{wu2023objectsdf++} &384x384 & \underline{0.0921}  & 0.7482 & 0.8105 & 0.0886 & 0.6168 & 0.8520 \\
PHYRECON \cite{ni2024phyrecon} &384x384 & \textbf{0.0792} & 0.7554 & \textbf{0.8254} & 0.0834 & 0.6301 & \textbf{0.8657} \\
GOC w/o OSC &384x384 & 0.1177  & \underline{0.7831} & 0.7979 & \textbf{0.0530} & \underline{0.7331} & 0.8324 \\
            &640x480 & 0.1444  & \textbf{0.7956} & \underline{0.8134}  & \underline{0.0556} & \textbf{0.8243} & \underline{0.8591} \\
\bottomrule
\end{tabular}%
}
\vspace*{-3mm}
\end{table}

Our method outperforms competing approaches in scene reconstruction metrics, achieving the lowest CD and the highest F-score, demonstrating superior geometric accuracy and completeness. In object reconstruction, Our method achieves the highest F-score, indicating strong object-level reconstruction quality. Overall, our method demonstrates strong performance in both object-compositional and full scene reconstruction.

\subsection{Ablation Study}
\label{subsec:exp_ablation_study}

\paragraph{Geometric Regularizations} 
To quantitatively analyze the effectiveness of the proposed regularizations, we on real scenes by comparing our full method to four variants in \tabref{tab:ablation_3d_gs_recon}. The ablation study demonstrates the significance of each regularization in our framework. Depth regularization ${\mathcal{L}_{d}}$ has the most substantial impact, with its removal causing severe performance drops, particularly in F-score, indicating its critical role in accurate geometry capture. Photo-consistency loss $\mathcal{L}_{pho}$ aids multi-view alignment, reducing inconsistencies across views, while denoising loss ${\mathcal{L}_{dn}}$ contributes to structural integrity and smoothness, as shown by its positive effect on F-score and NC. Geometric regularization $\mathcal{L}_{geo}$, though less impactful, helps fine-tune surface details. Together, these components enable robust, high-quality object-compositional reconstruction.

\begin{table}[ht]
\centering
\caption{Ablation Study of Scene Recon on Scannet \cite{dai2017scannet}}
\vspace*{-3mm}
\label{tab:ablation_3d_gs_recon}
\resizebox{0.7\linewidth}{!}{%
\begin{tabular}{lcccc}
\toprule
Method & CD $\downarrow$ & F-score $\uparrow$ & NC $\uparrow$ \\
\midrule
Full & 0.0556 & 0.8243 & \textbf{0.8591} \\
w/o ${\mathcal{L}_{dn}}$ & \textbf{0.0534} & \textbf{0.8430} & 0.8201 \\
w/o ${\mathcal{L}_{d}}$ & 0.1330 & 0.3483 & 0.7651 \\
w/o $\mathcal{L}_{pho}$ & 0.0568 & 0.8087 & 0.8591 \\
w/o $\mathcal{L}_{geo}$ & 0.0569 & 0.8246 & 0.8507 \\
\bottomrule
\end{tabular}%
}
\vspace*{-3mm}
\end{table}

\paragraph{OSC Model Object Reconstruction Quality}
We evaluated the OSC model on the ShapeNet \cite{chang2015shapenet} test set for surface reconstruction quality under complete point cloud inputs. Compared to state-of-the-art methods such as 3D2VS \cite{zhang20233dshape2vecset} and IF-Net \cite{chibane2020implicit}, OSC achieved the best reconstruction quality as shown in supplementary materials. This demonstrates that OSC is robust to different forms of point cloud inputs.
\paragraph{OSC Model Structure}
We experimented with different model architectures and evaluated the geometric accuracy of surface reconstruction on the ShapeNet test set, as shown in Table~\ref{tab:ablation_vae_model_structure}. Similar to MAE\cite{he2022masked}, we made the decoder lighter and concentrated more challenging learning tasks in the encoder, allowing for better adaptation to various potential downstream tasks. Ultimately, we selected the medium-sized model, OSC-M, with 101 million parameters, as it provides the optimal trade-off between performance and efficiency.

\begin{table}[ht]
\centering
\caption{Ablation study of the OSC model structure. Metrics were evaluated on the ShapeNet test set. “enc. Depth” refers to the number of self-attention and cross-attention layers in the encoder of the OSC model, while “dec.” refers to the decoder.}
\label{tab:ablation_vae_model_structure}
\vspace*{-3mm}
\resizebox{\linewidth}{!}{%
\begin{tabular}{lcccccc}
    \toprule
    Model & enc. Depth & dec. Depth & Params. & IoU↑ & CD↓ & F-score↑ \\
    \midrule
    OSC-S & 6 & 4 & 63 M & 0.969 & 0.018 & 0.983 \\
    OSC-M & 10 & 6 & 101 M & 0.975 & 0.018 & 0.987 \\
    OSC-L & 14 & 8 & 164 M & \textbf{0.976} & \textbf{0.017} & \textbf{0.990} \\
    \bottomrule
\end{tabular}%
}
\vspace*{-3mm}
\end{table}

\paragraph{OSC Model Training Data Augmentation}
In the OSC model training, we experimented with three different data preprocessing approaches: using complete point clouds, applying a random $50\%$ dropout to the point clouds, and simulating occlusions by filtering point clouds based on camera visibility. Results in Table~\ref{tab:ablation_vae_mask} from the synthetic dataset under Sparse Observation showed that simulating camera occlusions produced the best completion and reconstruction quality. Combining the second and third masking strategies did not yield further performance improvements. As a result, we adopted the occlusion-based data augmentation strategy for training.

\begin{table}[ht]
\centering
\caption{The impact of different masking strategies on completion quality. Metrics were evaluated on the synthetic dataset under Sparse Observation.}
\vspace*{-3mm}
\label{tab:ablation_vae_mask}
\resizebox{\linewidth}{!}{%
\begin{tabular}{lcccc}
\toprule
Method & Accuracy↓ & Completion↓ & CD↓ & F-score↑ \\
\midrule
w/o mask & 0.0115 & 0.0748 & 0.0432 & 0.8380 \\
Random drop & 0.0152 & \textbf{0.0567} & 0.0360 & 0.8822 \\
Visible mask & \textbf{0.0073} & 0.0575 & \textbf{0.0324} & \textbf{0.9033} \\
\bottomrule
\end{tabular}%
}
\vspace*{-3mm}
\end{table}

\section{Conclusion}
In this work, we present Gaussian Object Carver (GOC), a novel and efficient framework for object-compositional scene reconstruction. Compared to existing methods, GOC achieves more than 10 times efficiency, and generates watertight, separable object meshes, even in scenarios involving occlusion.
We introduce the zero-shot 3D Object Surface Completion (OSC) model, trained on a large-scale dataset, demonstrating generalizability for unseen surface completion at the object level.



%% file: sec/X_suppl.tex
\clearpage
\setcounter{page}{1}
\maketitlesupplementary

\section{OSC Reconstruction Quality on ShapeNet}
We evaluated the OSC model on the ShapeNet \cite{chang2015shapenet} test set to assess surface reconstruction quality using complete point-cloud inputs. Compared to state-of-the-art methods such as 3D2VS \cite{zhang20233dshape2vecset} and IF-Net \cite{chibane2020implicit}, OSC demonstrated superior performance across all metrics.

\begin{table}[ht]
\centering
\caption{Comparison of Single Object Surface Reconstruction Quality on the ShapeNet Test Set.}
\label{tab:vae_recons}
\vspace*{-3mm}
\resizebox{0.7\linewidth}{!}{%
\begin{tabular}{lccc}
    \toprule
    Model & IoU↑ & CD↓ & F-score↑ \\
    \midrule
    OccNet \cite{mescheder2019occupancy} & 0.825 & 0.072 & 0.858 \\
    ConvOccNet \cite{peng2020convolutional} & 0.888 & 0.052 & 0.933 \\
    IF-Net \cite{chibane2020implicit} & 0.934 & 0.041 & 0.967 \\
    3DILG \cite{zhang20223dilg} & 0.953 & 0.040 & 0.970 \\
    3D2VS \cite{zhang20233dshape2vecset} & 0.965 & 0.038 & 0.967 \\
    OSC & \textbf{0.975} & \textbf{0.018} & \textbf{0.987} \\
    \bottomrule
\end{tabular}%
}
\vspace*{-3mm}
\end{table}

Table. \ref{tab:vae_recons} provides a detailed comparison of single-object surface reconstruction methods on the ShapeNet test set. The OSC model outperformed prior approaches, achieving the highest Intersection over Union (IoU) at 0.975, the lowest Chamfer Distance (CD) at 0.018, and the highest F-score at 0.987. These results highlight OSC's robustness and effectiveness in reconstructing precise geometric surfaces.

The significant improvements in IoU, CD, and F-score metrics underline the model's ability to capture fine-grained geometric details and achieve accurate surface reconstructions. This establishes OSC as a leading approach for robust surface reconstruction, particularly when using complete point-cloud data.

\section{Additional Ablation Results for OSC Model}

\begin{figure}
    \vspace*{-3mm}
    \centering
    \includegraphics[width=\linewidth]{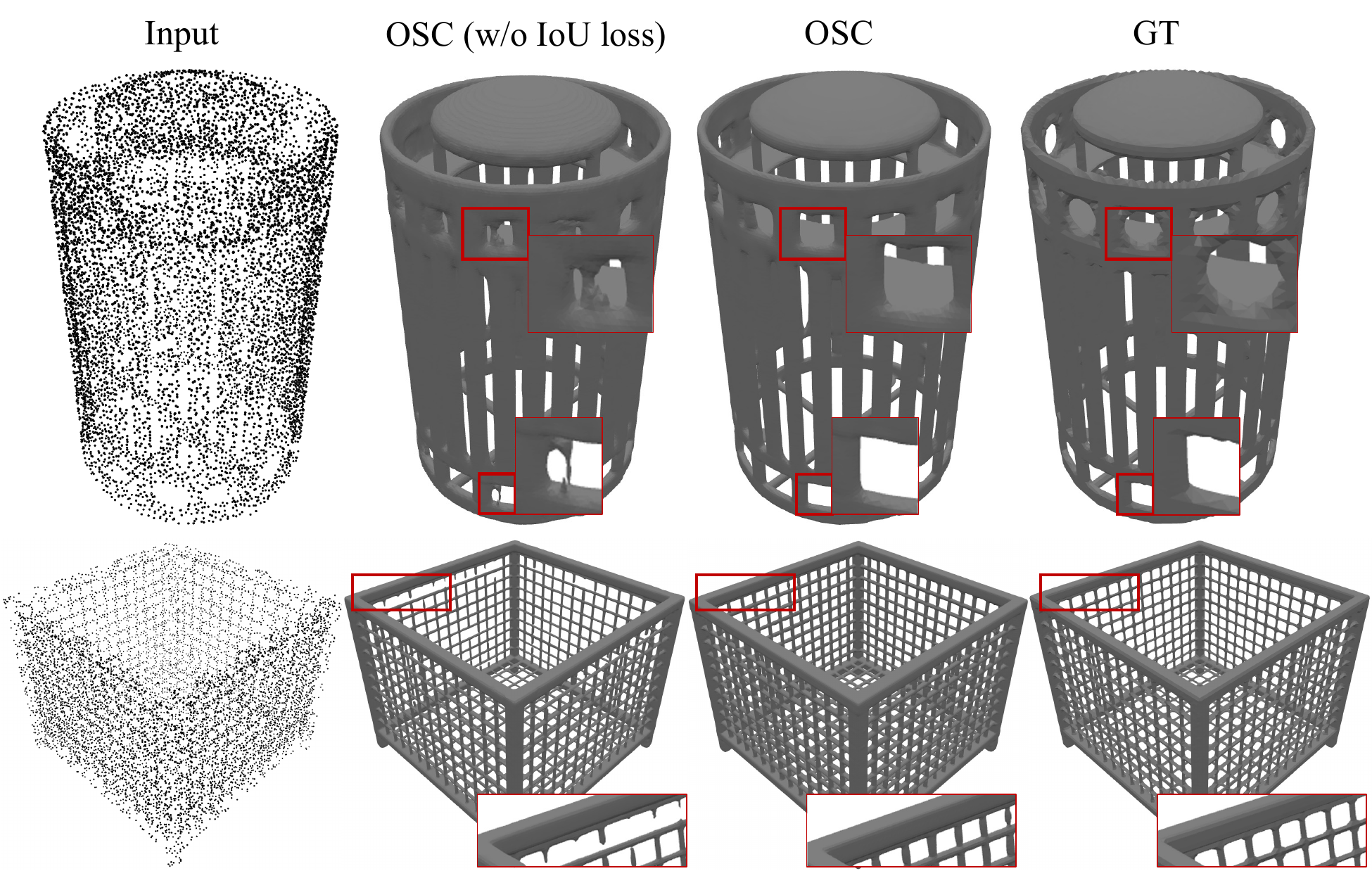}
    \caption{Qualitative comparison of reconstruction quality with (second column) and without $\mathcal{L}_{\text{IoU}}$ (third column). Using $\mathcal{L}_{\text{IoU}}$  aids OSC in establishing a well-defined isosurface threshold during the inference stage, resulting in clear and sharp mesh boundaries. Zoom in for details.}
    \label{fig:vae_ablation_iou}
    \vspace*{-3mm}
\end{figure}

\begin{figure}
    \vspace*{-3mm}
    \centering
    \includegraphics[width=\linewidth]{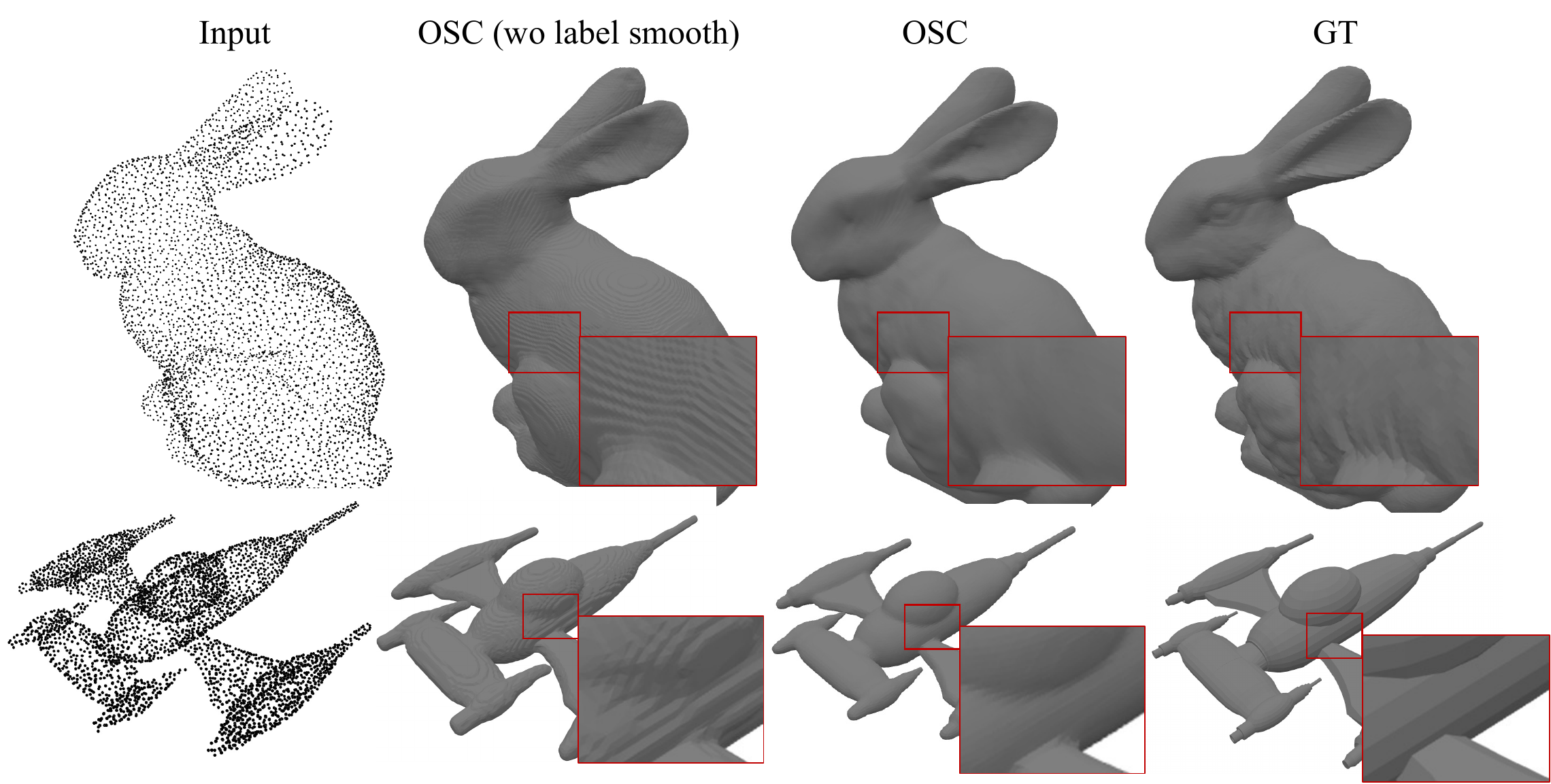}
    \caption{Qualitative comparison of reconstruction quality with (second column) and without label smoothing (third column). Label smoothing enhances the precision and smoothness of the reconstructed mesh surface, effectively reducing voxel-like artifacts on the surface. Zoom in for details.}
    \label{fig:vae_ablation_smooth}
    \vspace*{-5mm}
\end{figure}

\begin{figure*}[t]
\vspace*{-3mm}
\centering
\includegraphics[width=\textwidth]{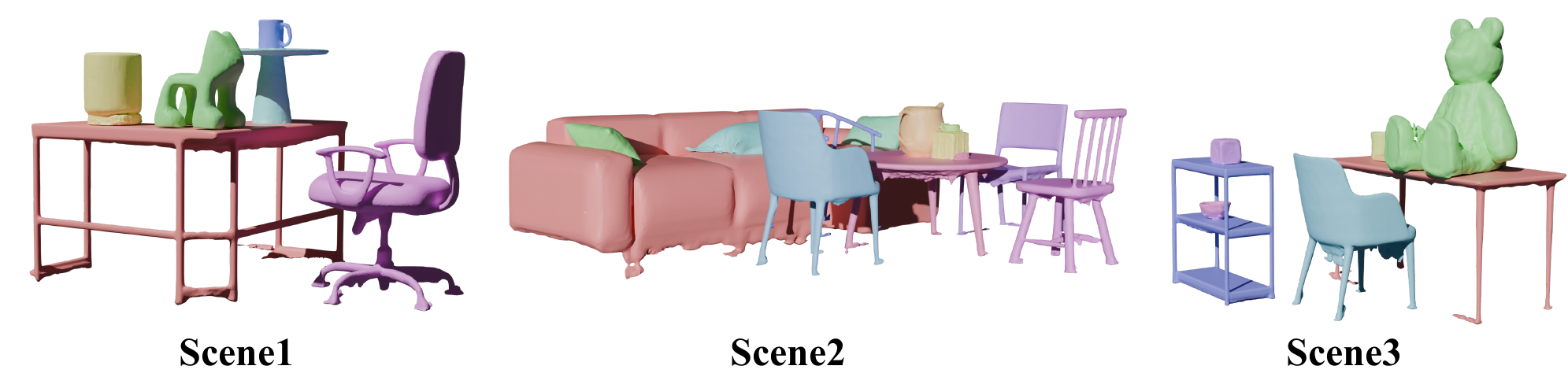}
\caption{Semantic mesh results of GOC on Synthetic Scenes}
\label{fig:fig2}
\vspace*{-3mm}
\end{figure*}

To evaluate the effectiveness of key components in the OSC model, we performed an ablation study with additional experiments focusing on $\mathcal{L}_{\text{IoU}}$ loss and label smoothing. The results, presented in Figures \ref{fig:vae_ablation_iou} and \ref{fig:vae_ablation_smooth} , reveal the significant impact of these components on reconstruction quality. Excluding $\mathcal{L}_{\text{IoU}}$ results in poorly defined mesh boundaries, highlighting its role in establishing accurate isosurface thresholds during inference. In contrast, including $\mathcal{L}_{\text{IoU}}$ ensures sharp and precise boundary delineation. Similarly, the absence of label smoothing leads to voxel-like artifacts that degrade surface quality, whereas its inclusion enhances smoothness and detail, producing refined and artifact-free meshes. These additional results confirm the critical contributions of $\mathcal{L}_{\text{IoU}}$ loss and label smoothing to the overall performance and robustness of the OSC model.

\section{Additional 3D GS Implementation Details}
\paragraph{Implementation Details}
Our code is built based on gsplat \cite{gsplat}and training strategy are consistent with \cite{3d_gs_mcmc}, because we observed that 3DGS \cite{kerbl20233d_3dgs} strategy is sensitive with initialization and hyperparameter settings. The training iterations for all scenes are set to 30,000. All experiments in this paper are conducted on Nvidia RTX 4090 GPU.

\paragraph{Mesh Exaction}  We start by rendering the depth for each training view and then apply the TSDF Fusion algorithm \cite{newcombe2011kinectfusion_tsdf} to construct the corresponding TSDF field. From this field, the mesh is subsequently extracted using the Marching Cubes algorithm \cite{lorensen1998marching_cube}

\paragraph{Depth Regularization}  For datasets with sensor-provided depth at scene scale, L1 loss can be directly used for supervision. However, in the absence of such sensor data, a monocular depth model is employed to generate a prior, albeit without actual depth measurements. The inherent scale ambiguity in monocular depth estimates must be addressed to align them with true scene geometry. To achieve this, we employ least squares optimization to refine both the scaling parameter ${k}$ and the offset parameter ${b}$ for each image. This ensures that the monocular depth estimates are consistent with the rendered depth in terms of scale:

\begin{equation} 
\hat{k}, \hat{b} = \underset{k, b}{\arg\min} \sum_{i,j} \left| \left( k \cdot \hat{D}_{i,j} + b \right) - D_{i,j} \right|_2^2, \label{eq:depth_proj} 
\end{equation}

where ${\hat{D}_{i,j}}$ and ${D_{i,j}}$ are the per-pixel depth values of the predicted and rendered depth maps, respectively. Once aligned, we apply the same loss function as used for sensor depth regularization.

\section{Additional results} 
Per-scene quantitative results of GOC on the Synthetic Scenes are reported in Fig. \ref{fig:fig2}. This process yields watertight and separable object meshes while preserving highly detailed features, enabling flexible scene rearrangement and object-level manipulation

\section{Limitation}
Currently, our approach supports geometry completion based solely on reconstructed point cloud data. It's simple and efficient but may struggle with complex object models due to ambiguity. In future work, we aim to integrate additional observations into the 3D model input, such as multiview CLIP features and texture information, leveraging multimodal data to achieve more accurate 3D completion and generation. This enhancement will enable a more robust integration of scene observations with data-driven priors.

%% file: main.bbl
\begin{thebibliography}{62}
\providecommand{\natexlab}[1]{#1}
\providecommand{\url}[1]{\texttt{#1}}
\expandafter\ifx\csname urlstyle\endcsname\relax
  \providecommand{\doi}[1]{doi: #1}\else
  \providecommand{\doi}{doi: \begingroup \urlstyle{rm}\Url}\fi

\bibitem[{Blender team}()]{Blender}
{Blender team}.
\newblock \emph{Blender: 3D modelling and rendering package}.
\newblock Available at \url{https://www.blender.org}.

\bibitem[{BlenderKit Team}()]{BlenderKit}
{BlenderKit Team}.
\newblock Blenderkit.
\newblock Online; accessed 14 November 2024.
\newblock Available at \url{https://www.blenderkit.com}.

\bibitem[Chang et~al.(2015)Chang, Funkhouser, Guibas, Hanrahan, Huang, Li, Savarese, Savva, Song, Su, et~al.]{chang2015shapenet}
Angel~X Chang, Thomas Funkhouser, Leonidas Guibas, Pat Hanrahan, Qixing Huang, Zimo Li, Silvio Savarese, Manolis Savva, Shuran Song, Hao Su, et~al.
\newblock Shapenet: An information-rich 3d model repository.
\newblock \emph{arXiv preprint arXiv:1512.03012}, 2015.

\bibitem[Chen et~al.(2024)Chen, Li, Ye, Wang, Xie, Zhai, Wang, Liu, Bao, and Zhang]{chen2024pgsr}
Danpeng Chen, Hai Li, Weicai Ye, Yifan Wang, Weijian Xie, Shangjin Zhai, Nan Wang, Haomin Liu, Hujun Bao, and Guofeng Zhang.
\newblock Pgsr: Planar-based gaussian splatting for efficient and high-fidelity surface reconstruction.
\newblock \emph{arXiv preprint arXiv:2406.06521}, 2024.

\bibitem[Chibane et~al.(2020)Chibane, Alldieck, and Pons-Moll]{chibane2020implicit}
Julian Chibane, Thiemo Alldieck, and Gerard Pons-Moll.
\newblock Implicit functions in feature space for 3d shape reconstruction and completion.
\newblock In \emph{Proceedings of the IEEE/CVF conference on computer vision and pattern recognition}, pages 6970--6981, 2020.

\bibitem[Chu et~al.(2024)Chu, Xie, Mo, Li, Nie{\ss}ner, Fu, and Jia]{chu2024diffcomplete}
Ruihang Chu, Enze Xie, Shentong Mo, Zhenguo Li, Matthias Nie{\ss}ner, Chi-Wing Fu, and Jiaya Jia.
\newblock Diffcomplete: Diffusion-based generative 3d shape completion.
\newblock \emph{Advances in Neural Information Processing Systems}, 36, 2024.

\bibitem[Dai et~al.(2017)Dai, Chang, Savva, Halber, Funkhouser, and Nie{\ss}ner]{dai2017scannet}
Angela Dai, Angel~X Chang, Manolis Savva, Maciej Halber, Thomas Funkhouser, and Matthias Nie{\ss}ner.
\newblock Scannet: Richly-annotated 3d reconstructions of indoor scenes.
\newblock In \emph{Proceedings of the IEEE conference on computer vision and pattern recognition}, pages 5828--5839, 2017.

\bibitem[Deitke et~al.(2023)Deitke, Schwenk, Salvador, Weihs, Michel, VanderBilt, Schmidt, Ehsani, Kembhavi, and Farhadi]{deitke2023objaverse}
Matt Deitke, Dustin Schwenk, Jordi Salvador, Luca Weihs, Oscar Michel, Eli VanderBilt, Ludwig Schmidt, Kiana Ehsani, Aniruddha Kembhavi, and Ali Farhadi.
\newblock Objaverse: A universe of annotated 3d objects.
\newblock In \emph{Proceedings of the IEEE/CVF Conference on Computer Vision and Pattern Recognition}, pages 13142--13153, 2023.

\bibitem[Fu et~al.(2022)Fu, Xu, Ong, and Tao]{Fu2022GeoNeus}
Qiancheng Fu, Qingshan Xu, Yew-Soon Ong, and Wenbing Tao.
\newblock Geo-neus: Geometry-consistent neural implicit surfaces learning for multi-view reconstruction.
\newblock \emph{Advances in Neural Information Processing Systems (NeurIPS)}, 2022.

\bibitem[Godard et~al.(2017)Godard, {Mac Aodha}, and Brostow]{monodepth17}
Cl{\'{e}}ment Godard, Oisin {Mac Aodha}, and Gabriel~J. Brostow.
\newblock Unsupervised monocular depth estimation with left-right consistency.
\newblock In \emph{CVPR}, 2017.

\bibitem[Gu{\'e}don and Lepetit(2024)]{guedon2024sugar}
Antoine Gu{\'e}don and Vincent Lepetit.
\newblock Sugar: Surface-aligned gaussian splatting for efficient 3d mesh reconstruction and high-quality mesh rendering.
\newblock In \emph{Proceedings of the IEEE/CVF Conference on Computer Vision and Pattern Recognition}, pages 5354--5363, 2024.

\bibitem[Hassena et~al.(2024)Hassena, Moon, Fujii, Yuen, Snavely, Marschner, and Hariharan]{hassena2024objectcarver}
Gemmechu Hassena, Jonathan Moon, Ryan Fujii, Andrew Yuen, Noah Snavely, Steve Marschner, and Bharath Hariharan.
\newblock Objectcarver: Semi-automatic segmentation, reconstruction and separation of 3d objects.
\newblock \emph{arXiv preprint arXiv:2407.19108}, 2024.

\bibitem[He et~al.(2022)He, Chen, Xie, Li, Doll{\'a}r, and Girshick]{he2022masked}
Kaiming He, Xinlei Chen, Saining Xie, Yanghao Li, Piotr Doll{\'a}r, and Ross Girshick.
\newblock Masked autoencoders are scalable vision learners.
\newblock In \emph{Proceedings of the IEEE/CVF conference on computer vision and pattern recognition}, pages 16000--16009, 2022.

\bibitem[Huang et~al.(2024)Huang, Yu, Chen, Geiger, and Gao]{huang20242d}
Binbin Huang, Zehao Yu, Anpei Chen, Andreas Geiger, and Shenghua Gao.
\newblock 2d gaussian splatting for geometrically accurate radiance fields.
\newblock In \emph{ACM SIGGRAPH 2024 Conference Papers}, pages 1--11, 2024.

\bibitem[Huang et~al.(2023)Huang, Gojcic, Atzmon, Litany, Fidler, and Williams]{huang2023neural}
Jiahui Huang, Zan Gojcic, Matan Atzmon, Or Litany, Sanja Fidler, and Francis Williams.
\newblock Neural kernel surface reconstruction.
\newblock In \emph{Proceedings of the IEEE/CVF Conference on Computer Vision and Pattern Recognition}, pages 4369--4379, 2023.

\bibitem[Kazhdan et~al.(2006)Kazhdan, Bolitho, and Hoppe]{kazhdan2006poisson}
Michael Kazhdan, Matthew Bolitho, and Hugues Hoppe.
\newblock Poisson surface reconstruction.
\newblock In \emph{Proceedings of the fourth Eurographics symposium on Geometry processing}, 2006.

\bibitem[Kerbl et~al.(2023)Kerbl, Kopanas, Leimk{\"u}hler, and Drettakis]{kerbl20233d_3dgs}
Bernhard Kerbl, Georgios Kopanas, Thomas Leimk{\"u}hler, and George Drettakis.
\newblock 3d gaussian splatting for real-time radiance field rendering.
\newblock \emph{ACM Trans. Graph.}, 42\penalty0 (4):\penalty0 139--1, 2023.

\bibitem[Kheradmand et~al.(2024)Kheradmand, Rebain, Sharma, Sun, Tseng, Isack, Kar, Tagliasacchi, and Yi]{3d_gs_mcmc}
Shakiba Kheradmand, Daniel Rebain, Gopal Sharma, Weiwei Sun, Jeff Tseng, Hossam Isack, Abhishek Kar, Andrea Tagliasacchi, and Kwang~Moo Yi.
\newblock 3d gaussian splatting as markov chain monte carlo.
\newblock \emph{arXiv preprint arXiv:2404.09591}, 2024.

\bibitem[Kingma and Welling(2013)]{Kingma2013AutoEncodingVB}
Diederik~P. Kingma and Max Welling.
\newblock Auto-encoding variational bayes.
\newblock \emph{CoRR}, abs/1312.6114, 2013.

\bibitem[Li et~al.(2023)Li, Lyu, Ding, Wang, Liao, and Liu]{li2023rico}
Zizhang Li, Xiaoyang Lyu, Yuanyuan Ding, Mengmeng Wang, Yiyi Liao, and Yong Liu.
\newblock Rico: Regularizing the unobservable for indoor compositional reconstruction.
\newblock In \emph{Proceedings of the IEEE/CVF International Conference on Computer Vision}, pages 17761--17771, 2023.

\bibitem[Liu et~al.(2024{\natexlab{a}})Liu, Shi, Chen, Zhang, Xu, Wei, Chen, Zeng, Gu, and Su]{liu2024one2}
Minghua Liu, Ruoxi Shi, Linghao Chen, Zhuoyang Zhang, Chao Xu, Xinyue Wei, Hansheng Chen, Chong Zeng, Jiayuan Gu, and Hao Su.
\newblock One-2-3-45++: Fast single image to 3d objects with consistent multi-view generation and 3d diffusion.
\newblock In \emph{Proceedings of the IEEE/CVF Conference on Computer Vision and Pattern Recognition}, pages 10072--10083, 2024{\natexlab{a}}.

\bibitem[Liu et~al.(2024{\natexlab{b}})Liu, Xu, Jin, Chen, Varma~T, Xu, and Su]{liu2024one}
Minghua Liu, Chao Xu, Haian Jin, Linghao Chen, Mukund Varma~T, Zexiang Xu, and Hao Su.
\newblock One-2-3-45: Any single image to 3d mesh in 45 seconds without per-shape optimization.
\newblock \emph{Advances in Neural Information Processing Systems}, 36, 2024{\natexlab{b}}.

\bibitem[Lorensen and Cline(1998)]{lorensen1998marching_cube}
William~E Lorensen and Harvey~E Cline.
\newblock Marching cubes: A high resolution 3d surface construction algorithm.
\newblock In \emph{Seminal graphics: pioneering efforts that shaped the field}, pages 347--353. 1998.

\bibitem[Mescheder et~al.(2019)Mescheder, Oechsle, Niemeyer, Nowozin, and Geiger]{mescheder2019occupancy}
Lars Mescheder, Michael Oechsle, Michael Niemeyer, Sebastian Nowozin, and Andreas Geiger.
\newblock Occupancy networks: Learning 3d reconstruction in function space.
\newblock In \emph{Proceedings of the IEEE/CVF conference on computer vision and pattern recognition}, pages 4460--4470, 2019.

\bibitem[Mildenhall et~al.(2021)Mildenhall, Srinivasan, Tancik, Barron, Ramamoorthi, and Ng]{mildenhall2021nerf}
Ben Mildenhall, Pratul~P Srinivasan, Matthew Tancik, Jonathan~T Barron, Ravi Ramamoorthi, and Ren Ng.
\newblock Nerf: Representing scenes as neural radiance fields for view synthesis.
\newblock \emph{Communications of the ACM}, 65\penalty0 (1):\penalty0 99--106, 2021.

\bibitem[M{\"u}ller et~al.(2019)M{\"u}ller, Kornblith, and Hinton]{muller2019does}
Rafael M{\"u}ller, Simon Kornblith, and Geoffrey~E Hinton.
\newblock When does label smoothing help?
\newblock \emph{Advances in neural information processing systems}, 32, 2019.

\bibitem[Newcombe et~al.(2011)Newcombe, Izadi, Hilliges, Molyneaux, Kim, Davison, Kohi, Shotton, Hodges, and Fitzgibbon]{newcombe2011kinectfusion_tsdf}
Richard~A Newcombe, Shahram Izadi, Otmar Hilliges, David Molyneaux, David Kim, Andrew~J Davison, Pushmeet Kohi, Jamie Shotton, Steve Hodges, and Andrew Fitzgibbon.
\newblock Kinectfusion: Real-time dense surface mapping and tracking.
\newblock In \emph{2011 10th IEEE international symposium on mixed and augmented reality}, pages 127--136. Ieee, 2011.

\bibitem[Ni et~al.(2024)Ni, Chen, Jing, Jiang, Wang, Dai, Zhu, Zhu, and Huang]{ni2024phyrecon}
Junfeng Ni, Yixin Chen, Bohan Jing, Nan Jiang, Bin Wang, Bo Dai, Yixin Zhu, Song-Chun Zhu, and Siyuan Huang.
\newblock Phyrecon: Physically plausible neural scene reconstruction.
\newblock \emph{arXiv preprint arXiv:2404.16666}, 2024.

\bibitem[Peng et~al.(2020)Peng, Niemeyer, Mescheder, Pollefeys, and Geiger]{peng2020convolutional}
Songyou Peng, Michael Niemeyer, Lars Mescheder, Marc Pollefeys, and Andreas Geiger.
\newblock Convolutional occupancy networks.
\newblock In \emph{Computer Vision--ECCV 2020: 16th European Conference, Glasgow, UK, August 23--28, 2020, Proceedings, Part III 16}, pages 523--540. Springer, 2020.

\bibitem[Petrov et~al.(2024)Petrov, Goyal, Thamizharasan, Kim, Gadelha, Averkiou, Chaudhuri, and Kalogerakis]{petrov2024gem3d}
Dmitry Petrov, Pradyumn Goyal, Vikas Thamizharasan, Vladimir Kim, Matheus Gadelha, Melinos Averkiou, Siddhartha Chaudhuri, and Evangelos Kalogerakis.
\newblock Gem3d: Generative medial abstractions for 3d shape synthesis.
\newblock In \emph{ACM SIGGRAPH 2024 Conference Papers}, pages 1--11, 2024.

\bibitem[Qi et~al.(2017)Qi, Yi, Su, and Guibas]{qi2017pointnet++}
Charles~Ruizhongtai Qi, Li Yi, Hao Su, and Leonidas~J Guibas.
\newblock Pointnet++: Deep hierarchical feature learning on point sets in a metric space.
\newblock \emph{Advances in neural information processing systems}, 30, 2017.

\bibitem[Qu et~al.(2024)Qu, Dai, Li, Lin, Cao, Zhang, and Ji]{qu2024goi}
Yansong Qu, Shaohui Dai, Xinyang Li, Jianghang Lin, Liujuan Cao, Shengchuan Zhang, and Rongrong Ji.
\newblock Goi: Find 3d gaussians of interest with an optimizable open-vocabulary semantic-space hyperplane.
\newblock In \emph{Proceedings of the 32nd ACM International Conference on Multimedia}, pages 5328--5337, 2024.

\bibitem[Rao et~al.(2022)Rao, Nie, and Dai]{rao2022patchcomplete}
Yuchen Rao, Yinyu Nie, and Angela Dai.
\newblock Patchcomplete: Learning multi-resolution patch priors for 3d shape completion on unseen categories.
\newblock \emph{Advances in Neural Information Processing Systems}, 35:\penalty0 34436--34450, 2022.

\bibitem[Shu et~al.(2019)Shu, Park, and Kwon]{shu20193d}
Dong~Wook Shu, Sung~Woo Park, and Junseok Kwon.
\newblock 3d point cloud generative adversarial network based on tree structured graph convolutions.
\newblock In \emph{Proceedings of the IEEE/CVF international conference on computer vision}, pages 3859--3868, 2019.

\bibitem[Tonderski et~al.(2024)Tonderski, Lindstr{\"o}m, Hess, Ljungbergh, Svensson, and Petersson]{tonderski2024neurad}
Adam Tonderski, Carl Lindstr{\"o}m, Georg Hess, William Ljungbergh, Lennart Svensson, and Christoffer Petersson.
\newblock Neurad: Neural rendering for autonomous driving.
\newblock In \emph{Proceedings of the IEEE/CVF Conference on Computer Vision and Pattern Recognition}, pages 14895--14904, 2024.

\bibitem[Torne et~al.(2024)Torne, Simeonov, Li, Chan, Chen, Gupta, and Agrawal]{torne2024reconciling}
Marcel Torne, Anthony Simeonov, Zechu Li, April Chan, Tao Chen, Abhishek Gupta, and Pulkit Agrawal.
\newblock Reconciling reality through simulation: A real-to-sim-to-real approach for robust manipulation.
\newblock \emph{arXiv preprint arXiv:2403.03949}, 2024.

\bibitem[Turkulainen et~al.(2024)Turkulainen, Ren, Melekhov, Seiskari, Rahtu, and Kannala]{turkulainen2024dn}
Matias Turkulainen, Xuqian Ren, Iaroslav Melekhov, Otto Seiskari, Esa Rahtu, and Juho Kannala.
\newblock Dn-splatter: Depth and normal priors for gaussian splatting and meshing.
\newblock \emph{arXiv preprint arXiv:2403.17822}, 2024.

\bibitem[Wang et~al.(2022)Wang, Wang, Long, Theobalt, Komura, Liu, and Wang]{wang2022neuris}
Jiepeng Wang, Peng Wang, Xiaoxiao Long, Christian Theobalt, Taku Komura, Lingjie Liu, and Wenping Wang.
\newblock Neuris: Neural reconstruction of indoor scenes using normal priors.
\newblock In \emph{European Conference on Computer Vision}, pages 139--155. Springer, 2022.

\bibitem[Wang et~al.(2021)Wang, Liu, Liu, Theobalt, Komura, and Wang]{wang2021neus}
Peng Wang, Lingjie Liu, Yuan Liu, Christian Theobalt, Taku Komura, and Wenping Wang.
\newblock Neus: Learning neural implicit surfaces by volume rendering for multi-view reconstruction.
\newblock \emph{arXiv preprint arXiv:2106.10689}, 2021.

\bibitem[WE(1987)]{we1987marching}
LORENSEN WE.
\newblock Marching cubes: A high resolution 3d surface construction algorithm.
\newblock \emph{Computer graphics}, 21\penalty0 (1):\penalty0 7--12, 1987.

\bibitem[Williams et~al.(2021)Williams, Trager, Bruna, and Zorin]{williams2021neural}
Francis Williams, Matthew Trager, Joan Bruna, and Denis Zorin.
\newblock Neural splines: Fitting 3d surfaces with infinitely-wide neural networks.
\newblock In \emph{Proceedings of the IEEE/CVF Conference on Computer Vision and Pattern Recognition}, pages 9949--9958, 2021.

\bibitem[Williams et~al.(2022)Williams, Gojcic, Khamis, Zorin, Bruna, Fidler, and Litany]{williams2022neural}
Francis Williams, Zan Gojcic, Sameh Khamis, Denis Zorin, Joan Bruna, Sanja Fidler, and Or Litany.
\newblock Neural fields as learnable kernels for 3d reconstruction.
\newblock In \emph{Proceedings of the IEEE/CVF Conference on Computer Vision and Pattern Recognition}, pages 18500--18510, 2022.

\bibitem[Wu et~al.(2024{\natexlab{a}})Wu, Liu, Cai, Yan, Wang, Hu, Duan, and Ma]{wu2024unique3d}
Kailu Wu, Fangfu Liu, Zhihan Cai, Runjie Yan, Hanyang Wang, Yating Hu, Yueqi Duan, and Kaisheng Ma.
\newblock Unique3d: High-quality and efficient 3d mesh generation from a single image.
\newblock \emph{arXiv preprint arXiv:2405.20343}, 2024{\natexlab{a}}.

\bibitem[Wu et~al.(2022)Wu, Liu, Chen, Li, Zheng, Cai, and Zheng]{wu2022object}
Qianyi Wu, Xian Liu, Yuedong Chen, Kejie Li, Chuanxia Zheng, Jianfei Cai, and Jianmin Zheng.
\newblock Object-compositional neural implicit surfaces.
\newblock In \emph{European Conference on Computer Vision}, pages 197--213. Springer, 2022.

\bibitem[Wu et~al.(2023)Wu, Wang, Li, Zheng, and Cai]{wu2023objectsdf++}
Qianyi Wu, Kaisiyuan Wang, Kejie Li, Jianmin Zheng, and Jianfei Cai.
\newblock Objectsdf++: Improved object-compositional neural implicit surfaces.
\newblock In \emph{Proceedings of the IEEE/CVF International Conference on Computer Vision}, pages 21764--21774, 2023.

\bibitem[Wu et~al.(2025)Wu, Zheng, Wu, and Cham]{wu2025clusteringsdf}
Tianhao Wu, Chuanxia Zheng, Qianyi Wu, and Tat-Jen Cham.
\newblock Clusteringsdf: Self-organized neural implicit surfaces for 3d decomposition.
\newblock In \emph{European Conference on Computer Vision}, pages 255--272. Springer, 2025.

\bibitem[Wu et~al.(2024{\natexlab{b}})Wu, Meng, Li, Wu, Shi, Cheng, Zhao, Feng, Ding, Wang, et~al.]{wu2024opengaussian}
Yanmin Wu, Jiarui Meng, Haijie Li, Chenming Wu, Yahao Shi, Xinhua Cheng, Chen Zhao, Haocheng Feng, Errui Ding, Jingdong Wang, et~al.
\newblock Opengaussian: Towards point-level 3d gaussian-based open vocabulary understanding.
\newblock \emph{arXiv preprint arXiv:2406.02058}, 2024{\natexlab{b}}.

\bibitem[Xie et~al.(2020)Xie, Yao, Zhou, Mao, Zhang, and Sun]{xie2020grnet}
Haozhe Xie, Hongxun Yao, Shangchen Zhou, Jiageng Mao, Shengping Zhang, and Wenxiu Sun.
\newblock Grnet: Gridding residual network for dense point cloud completion.
\newblock In \emph{European conference on computer vision}, pages 365--381. Springer, 2020.

\bibitem[Xu et~al.(2024)Xu, Cheng, Gao, Wang, Gao, and Shan]{xu2024instantmesh}
Jiale Xu, Weihao Cheng, Yiming Gao, Xintao Wang, Shenghua Gao, and Ying Shan.
\newblock Instantmesh: Efficient 3d mesh generation from a single image with sparse-view large reconstruction models.
\newblock \emph{arXiv preprint arXiv:2404.07191}, 2024.

\bibitem[Yan et~al.(2022)Yan, Lin, Mitra, Lischinski, Cohen-Or, and Huang]{yan2022shapeformer}
Xingguang Yan, Liqiang Lin, Niloy~J Mitra, Dani Lischinski, Daniel Cohen-Or, and Hui Huang.
\newblock Shapeformer: Transformer-based shape completion via sparse representation.
\newblock In \emph{Proceedings of the IEEE/CVF Conference on Computer Vision and Pattern Recognition}, pages 6239--6249, 2022.

\bibitem[Yang et~al.(2023)Yang, Chen, Wang, Manivasagam, Ma, Yang, and Urtasun]{yang2023unisim}
Ze Yang, Yun Chen, Jingkang Wang, Sivabalan Manivasagam, Wei-Chiu Ma, Anqi~Joyce Yang, and Raquel Urtasun.
\newblock Unisim: A neural closed-loop sensor simulator.
\newblock In \emph{Proceedings of the IEEE/CVF Conference on Computer Vision and Pattern Recognition}, pages 1389--1399, 2023.

\bibitem[Ye et~al.(2023)Ye, Danelljan, Yu, and Ke]{ye2023gaussian}
Mingqiao Ye, Martin Danelljan, Fisher Yu, and Lei Ke.
\newblock Gaussian grouping: Segment and edit anything in 3d scenes.
\newblock \emph{arXiv preprint arXiv:2312.00732}, 2023.

\bibitem[Ye et~al.(2024{\natexlab{a}})Ye, Danelljan, Yu, and Ke]{gaussian_grouping}
Mingqiao Ye, Martin Danelljan, Fisher Yu, and Lei Ke.
\newblock Gaussian grouping: Segment and edit anything in 3d scenes.
\newblock In \emph{ECCV}, 2024{\natexlab{a}}.

\bibitem[Ye et~al.(2024{\natexlab{b}})Ye, Li, Kerr, Turkulainen, Yi, Pan, Seiskari, Ye, Hu, Tancik, and Kanazawa]{gsplat}
Vickie Ye, Ruilong Li, Justin Kerr, Matias Turkulainen, Brent Yi, Zhuoyang Pan, Otto Seiskari, Jianbo Ye, Jeffrey Hu, Matthew Tancik, and Angjoo Kanazawa.
\newblock gsplat: An open-source library for {Gaussian} splatting.
\newblock \emph{arXiv preprint arXiv:2409.06765}, 2024{\natexlab{b}}.

\bibitem[Yu et~al.(2022)Yu, Peng, Niemeyer, Sattler, and Geiger]{yu2022monosdf}
Zehao Yu, Songyou Peng, Michael Niemeyer, Torsten Sattler, and Andreas Geiger.
\newblock Monosdf: Exploring monocular geometric cues for neural implicit surface reconstruction.
\newblock \emph{Advances in neural information processing systems}, 35:\penalty0 25018--25032, 2022.

\bibitem[Yuan et~al.(2018)Yuan, Khot, Held, Mertz, and Hebert]{yuan2018pcn}
Wentao Yuan, Tejas Khot, David Held, Christoph Mertz, and Martial Hebert.
\newblock Pcn: Point completion network.
\newblock In \emph{2018 international conference on 3D vision (3DV)}, pages 728--737. IEEE, 2018.

\bibitem[Zhang et~al.(2022)Zhang, Nie{\ss}ner, and Wonka]{zhang20223dilg}
Biao Zhang, Matthias Nie{\ss}ner, and Peter Wonka.
\newblock 3dilg: Irregular latent grids for 3d generative modeling.
\newblock \emph{Advances in Neural Information Processing Systems}, 35:\penalty0 21871--21885, 2022.

\bibitem[Zhang et~al.(2023)Zhang, Tang, Niessner, and Wonka]{zhang20233dshape2vecset}
Biao Zhang, Jiapeng Tang, Matthias Niessner, and Peter Wonka.
\newblock 3dshape2vecset: A 3d shape representation for neural fields and generative diffusion models.
\newblock \emph{ACM Transactions on Graphics (TOG)}, 42\penalty0 (4):\penalty0 1--16, 2023.

\bibitem[Zhang et~al.(2024)Zhang, Wang, Zhang, Qiu, Pang, Jiang, Yang, Xu, and Yu]{zhang2024clay}
Longwen Zhang, Ziyu Wang, Qixuan Zhang, Qiwei Qiu, Anqi Pang, Haoran Jiang, Wei Yang, Lan Xu, and Jingyi Yu.
\newblock Clay: A controllable large-scale generative model for creating high-quality 3d assets.
\newblock \emph{ACM Transactions on Graphics (TOG)}, 43\penalty0 (4):\penalty0 1--20, 2024.

\bibitem[Zhou et~al.(2018)Zhou, Park, and Koltun]{zhou2018open3d}
Qian-Yi Zhou, Jaesik Park, and Vladlen Koltun.
\newblock Open3d: A modern library for 3d data processing.
\newblock \emph{arXiv preprint arXiv:1801.09847}, 2018.

\bibitem[Zhou et~al.(2024)Zhou, Chang, Jiang, Fan, Zhu, Xu, Chari, You, Wang, and Kadambi]{zhou2024feature}
Shijie Zhou, Haoran Chang, Sicheng Jiang, Zhiwen Fan, Zehao Zhu, Dejia Xu, Pradyumna Chari, Suya You, Zhangyang Wang, and Achuta Kadambi.
\newblock Feature 3dgs: Supercharging 3d gaussian splatting to enable distilled feature fields.
\newblock In \emph{Proceedings of the IEEE/CVF Conference on Computer Vision and Pattern Recognition}, pages 21676--21685, 2024.

\bibitem[Zuo et~al.(2024)Zuo, Gu, Dong, Zhao, Yuan, Qiu, Bo, and Dong]{zuo2024sparse3d}
Qi Zuo, Xiaodong Gu, Yuan Dong, Zhengyi Zhao, Weihao Yuan, Lingteng Qiu, Liefeng Bo, and Zilong Dong.
\newblock High-fidelity 3d textured shapes generation by sparse encoding and adversarial decoding.
\newblock In \emph{European Conference on Computer Vision}, 2024.

\end{thebibliography}
